\pdfoutput=1
\documentclass[11pt,a4paper]{article}
\usepackage[hyperref]{emnlp2020}
\usepackage{times}
\usepackage{latexsym}
\usepackage{amsmath}
\usepackage{amsfonts}
\usepackage{amssymb}
\usepackage{latexsym}
\usepackage{booktabs}
\usepackage{color} 
\usepackage{setspace}
\usepackage{multirow}
\usepackage{url}
\usepackage{graphicx} 
\usepackage{subfigure}
\usepackage{algorithm}
\usepackage{algorithmic}
\usepackage{microtype}

\aclfinalcopy % Uncomment this line for the final submission
\title{Parsing All: Syntax and Semantics, Dependencies and Spans}

\author{Junru Zhou$^{1,2,3}$ , Zuchao Li $^{1,2,3}$,  Hai Zhao$^{1,2,3}$\thanks{$\ $  Corresponding author. This paper was partially supported by National Key Research and Development Program
of China (No. 2017YFB0304100), Key Projects of National Natural Science Foundation of China (U1836222 and
61733011), Huawei-SJTU long term AI project, Cutting-edge
Machine reading comprehension and language model.} \\
$^{1}$Department of Computer Science and Engineering, Shanghai Jiao Tong University \\
    $^{2}$Key Laboratory of Shanghai Education Commission for Intelligent Interaction \\ and Cognitive Engineering, Shanghai Jiao Tong University, Shanghai, China\\
    $^{3}$MoE Key Lab of Artificial Intelligence, AI Institute, Shanghai Jiao Tong University \\
  {\tt \{zhoujunru,charlee\}@sjtu.edu.cn, zhaohai@cs.sjtu.edu.cn}}

\date{}

\begin{document}
\maketitle

\begin{abstract}
Both syntactic and semantic structures are key linguistic contextual clues, in which parsing the latter has been well shown beneficial from parsing the former. However, few works ever made an attempt to let semantic parsing help syntactic parsing. As linguistic representation formalisms, both syntax and semantics may be represented in either span (constituent/phrase) or dependency, on both of which joint learning was also seldom explored.
In this paper, we propose a novel joint model of syntactic and semantic parsing on both span and dependency representations, which incorporates syntactic information effectively in the encoder of neural network and benefits from two representation formalisms in a uniform way.
The experiments show that semantics and syntax can benefit each other by optimizing joint objectives.
Our single model achieves new state-of-the-art or competitive results on both span and dependency semantic parsing on Propbank benchmarks and both dependency and constituent syntactic parsing on Penn Treebank.
 
\end{abstract}

\section{Introduction}

This work makes the first attempt to fill the gaps on syntactic and semantic parsing from jointly considering its representation forms and their linguistic processing layers. First, both span (constituent) and dependency are effective formal representations for both semantics and syntax, which have been well studied and discussed from both linguistic and computational perspective, though few works comprehensively considered the impact of either/both representation styles over the respective parsing  \cite{chomsky1981lectures,Li-aaai-19}. 
% Leeman2001Bresnan, johansson-nugues-2008-dependency-based, Li-aaai-19}.
% As show in Figure \ref{fig1}, the relationship between dependency and span formal, semantic role and syntax,
% while dependency SRL annotates the syntactic heads of arguments,
% span-style annotates entire argument spans.
Second, as semantics is usually considered as a higher layer of linguistics over syntax, most previous studies focus on how the latter helps the former. Though there comes a trend that syntactic clues show less impact on enhancing semantic parsing since neural models were introduced \cite{marcheggiani-titov-2017-encoding}. 
In fact, recent works \cite{he-etal-2017-deep,marcheggiani-etal-2017-simple} propose
syntax-agnostic models for semantic parsing and achieve competitive and even state-of-the-art results. However, semantics may not only benefit from syntax which has been well known, but syntax may also benefit from semantics, which is an obvious gap in explicit linguistic structure parsing and few attempts were ever reported.
To our best knowledge, few previous works focus on the relationship between syntax and semantic which only based on dependency structure \cite{swayamdipta-etal-2016-greedy,henderson-etal-2013-multilingual,shi-etal-2016-exploiting}. 
% For example, \cite{shi-etal-2016-exploiting} made a brief attempt on Chinese Semantic Treebank %\cite{qiu-2016} 
% to show the mutual benefits between dependency syntax and semantic roles.

% , which seems to be in conflict with
% the belief that syntactic information is an absolutely
% necessary prerequisite for high-performance
% SRL \cite{gildea-palmer-2002-necessity}.

% The SRL in our model learns in end-to-end way with a uniform representation and jointly predicts all predicates
% and arguments, especially including multi-predicates identification which is ignored in most of previous works.

To fill such a gap, in this work, we further exploit both strengths of the span and dependency representation of both semantic role labeling (SRL)
%lewis-etal-2015-joint
\cite{strubell-etal-2018-linguistically} and syntax, and propose a joint model\footnote{Our code : https://github.com/DoodleJZ/ParsingAll.} with multi-task learning in a balanced mode which improves both semantic and syntactic parsing. 
Moreover, in our model, semantics is learned in an end-to-end way with a uniform representation and syntactic parsing is represented as a $joint$ $span$ structure \cite{zhou-zhao-2019-head} relating to head-driven phrase structure grammar (HPSG) \cite{pollard1994head} which can incorporate both head and phrase information of dependency and constituent syntactic parsing.

We verify the effectiveness and applicability
of the proposed model on Propbank semantic parsing \footnote{It is also called semantic role labeling (SRL) for the semantic parsing task over the Propbank.} in both span style (CoNLL-2005) \cite{carreras-marquez-2005-introduction} and dependency style (CoNLL-2009) \cite{hajic-etal-2009-conll} and Penn Treebank (PTB) \cite{MarcusJ93-2004} for both constituent and dependency syntactic parsing.
% Additionally, we introduce a spacial label for SRL and syntactic parsing to take advantage of the relationship between dependency and span structure formulization, which enables our model to use a single decoder to implement SRL and syntactic parsing as HPSG parsing.
Our empirical results show that semantics and syntax can indeed benefit each other, and our single model reaches new state-of-the-art or competitive performance for all four tasks: span and dependency SRL, constituent and dependency syntactic parsing.

\section{Structure Representation}

In this section, we introduce a preprocessing method to handle span and dependency representation, which have strong inherent linguistic relation for both syntax and semantics.

For syntactic representation, 
we use a formal structure called $joint$ $span$ following \cite{zhou-zhao-2019-head} to cover both constituent and head information of syntactic tree based on HPSG which is a highly lexicalized, constraint-based grammar \cite{pollard1994head}.
For semantic (SRL) representation, we propose a unified structure to simplify the training process and employ SRL constraints for span arguments to enforce exact inference.

\begin{figure}[t!]
    \centering
    \subfigure[Constituent and dependency.]{
        \label{Fig.sub.1}
        \includegraphics[width=2in]{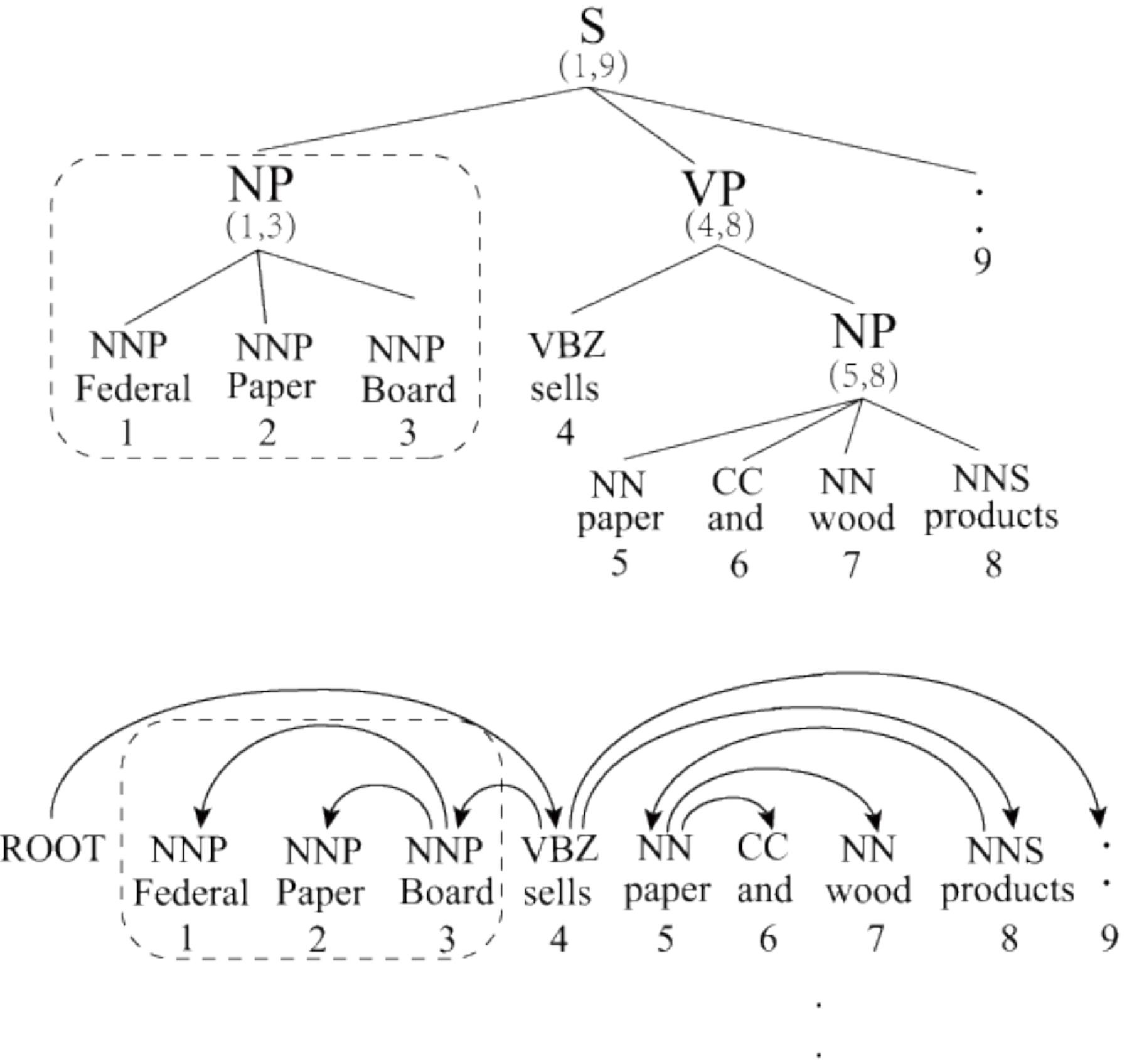}
    }
    \subfigure[Joint span structure.]{
        \label{Fig.sub.3}
        \includegraphics[width=2in]{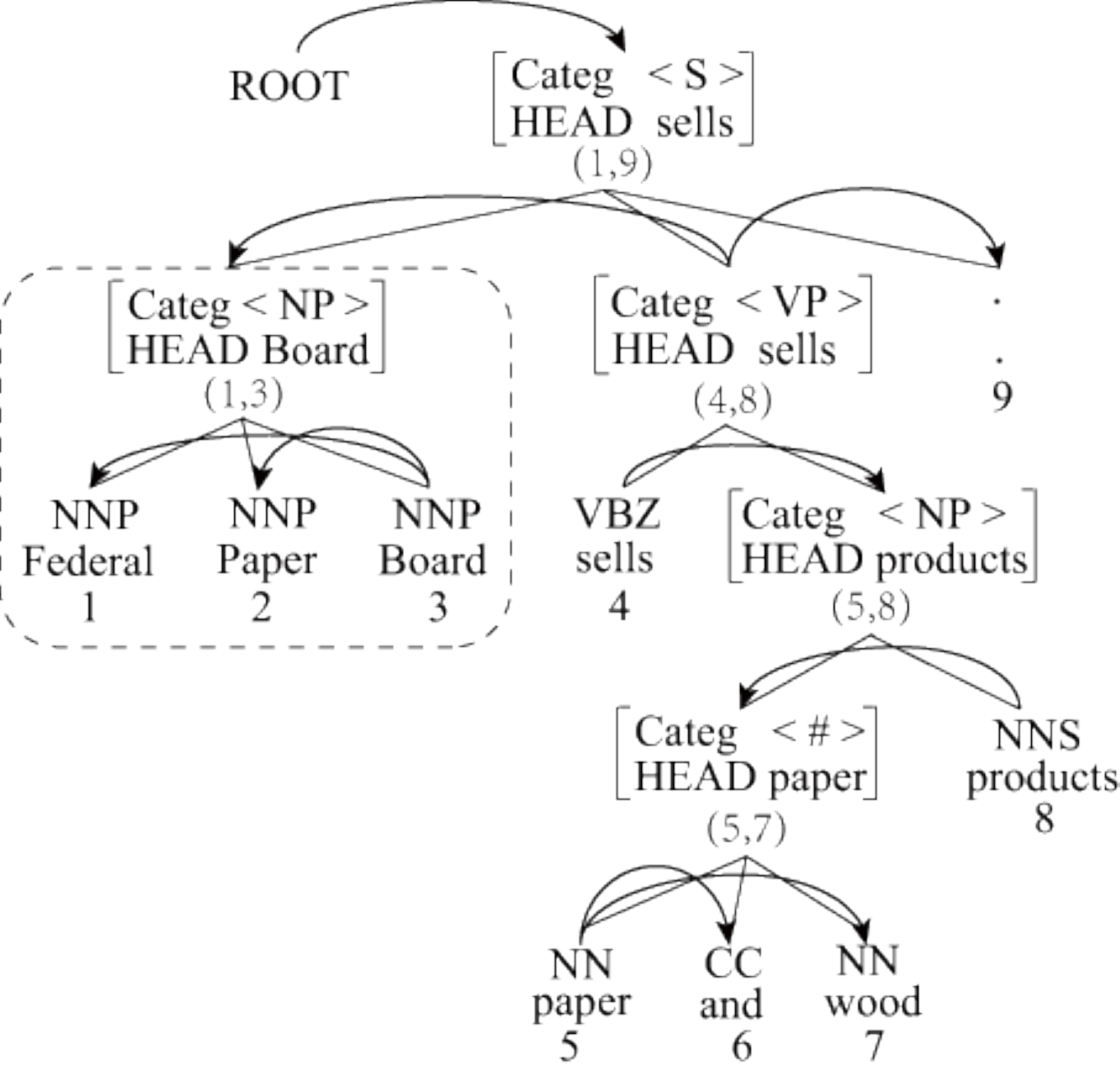}
    }
    \caption{Constituent, dependency, and joint span structures from \cite{zhou-zhao-2019-head}, which is indexed from 1 to 9 and assigned interval range for each node. The dotted box represents the same part. The special category $\#$ is assigned to divide the phrase with multiple heads. Joint span structure contains constitute phrase and dependency arc. \textit{Categ} in each node represents the category of each constituent, and \textit{HEAD} indicates the head word.}
    \label{fig2}
\end{figure}

\begin{figure*}[t!]
    \centering
    \includegraphics[width=6in]{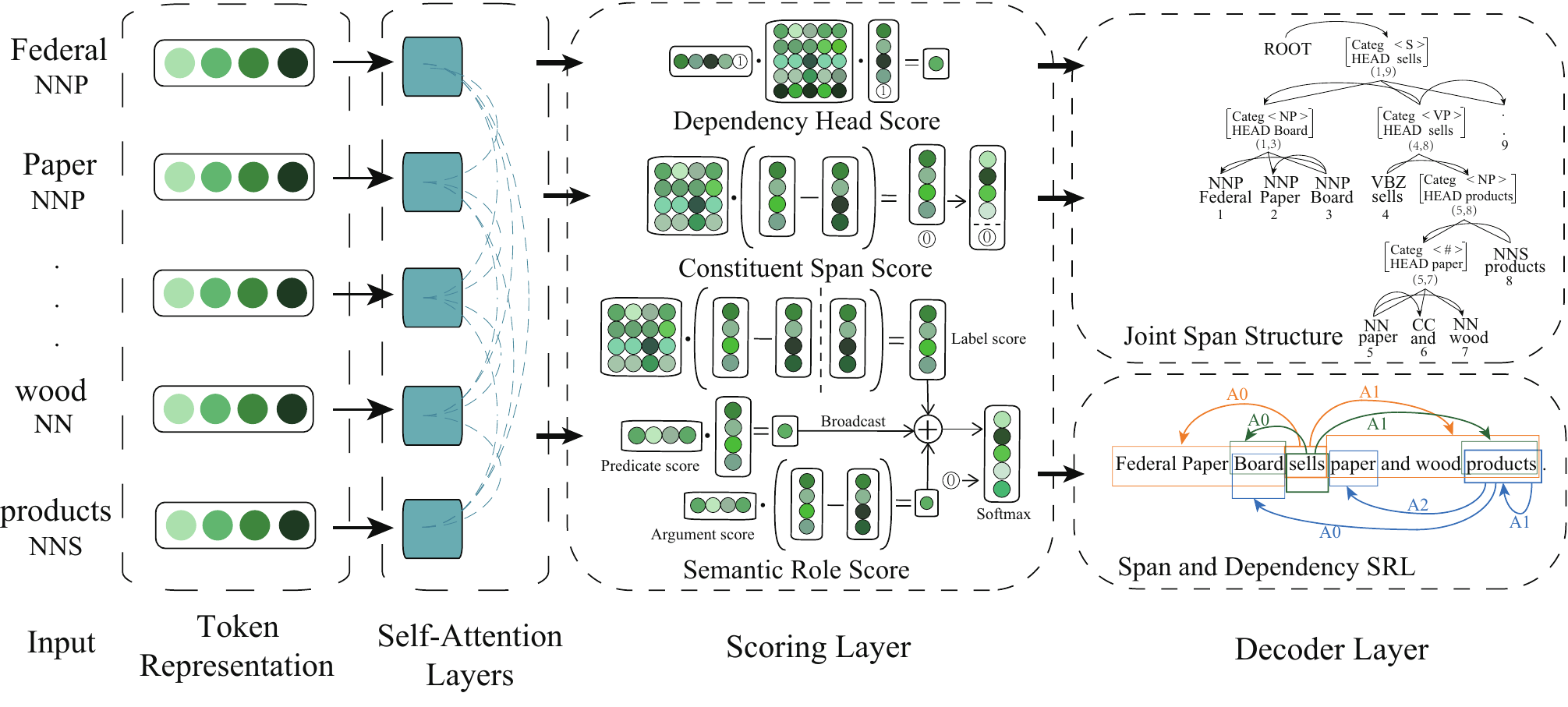}
    
    \caption{The framework of our joint parsing model.}
    \label{fig3}
\end{figure*}
\subsection{Syntactic Representation}

The $joint$ $span$ structure which is related to the HEAD FEATURE PRINCIPLE (HFP) of HPSG \cite{pollard1994head} consists of all its children phrases in the constituent tree and all dependency arcs between the head and children in the dependency tree.
% We formally define a structure following the HEAD FEATURE PRINCIPLE (HFP) of HPSG \cite{pollard1994head} called joint span which consists of all its children phrases in the constituent tree and all dependency arcs between heads of children phrases in the dependency tree. 

For example, in the constituent tree of Figure \ref{Fig.sub.1}, \textit{Federal Paper Board} is a phrase $(1,3)$ assigned with category NP and in dependency tree, \textit{Board} is parent of \textit{Federal} and \textit{Paper}, thus in our joint span structure, the head of phrase $(1,3)$ is \textit{Board}. 
The node S$_{H}$(1, 9) in Figure \ref{Fig.sub.3} as a joint span is:
S$_{H}$(1, 9) = \{ S$_{H}$(1, 3) , S$_{H}$(4, 8) , S$_{H}$(9, 9), $l$(1, 9, $<$S$>$) , $d$($Board$, $sells$) , $d$(., $sells$) \},
where $l$($i$, $j$, $<$S$>$) denotes category of span
($i$, $j$) with category S and $d$($r$, $h$) indicates the dependency
between the word $r$ and its parent $h$. 
At last, the entire syntactic tree $T$ being a joint span can be represented as:

$S_H$($T$) = \{$S_H$(1, 9), $d$($sells$, $root$)\}\footnote{For dependency label of each word, we  train a separated multi-class classifier simultaneously with the parser by optimizing the sum of their objectives.}.

Following most of the recent work, we apply the PTB-SD representation converted by version 3.3.0 of the Stanford parser.
However, this dependency representation results in around 1\% of phrases containing two or three head words. 
As shown in Figure \ref{Fig.sub.1}, the phrase (5,8) assigned with a category NP contains 2 head words of \textit{paper} and \textit{products} in dependency tree.
To deal with the problem, we introduce a special category $\#$ to divide the phrase with multiple heads to meet the criterion that there is only one head word for each phrase. After this conversion, only 50 heads are errors in PTB.
% \footnote{Example error in the training dataset:
% Constituent syntax:
% (SBARQ (WHNP (WP What)) (SQ (VBZ is) (NP (JJ greatness))) (. ?))
% Dependency head:
% [0, 1, 1, 1]
% The word “What” is the father of “is” and “greatness”, thus the phrase (SQ (VBZ is) (NP (JJ greatness))) has not head word. This error is not belong to multiple dependency head errors, thus we do not make any modification to the constituent or dependency head (we directly use this constituent tree and dependency to optimize the loss).
% }

% The node S$_{H}$(1, 9) in Figure \ref{Fig.sub.3} as a joint span is:
% S$_{H}$(1, 9) = \{ S$_{H}$(1, 3) , S$_{H}$(4, 8) , S$_{H}$(9, 9), $l$(1, 9) , $d$($Board$, $sells$) , $d$(., $sells$) \},

% \noindent where $l$($i$, $j$) denotes category of span
% ($i$, $j$) and $d$($r$, $h$) indicates the dependency
% between the word $r$ and its parent $h$. 

% At last, the entire syntactic tree $T$ being a joint span can be represented as:

% $S_H$($T$) = \{$S_H$(1, 9), $d$($sells$, $root$)\}.

Moreover, to simplify the syntactic parsing algorithm, we add a special empty category $\O$ to spans to binarize the n-ary nodes and apply a unary atomic category to deal with the nodes of the unary chain, which is popularly adopted in constituent syntactic parsing \cite{SternD17b,Gaddy}.
% As all constituent and head information has been formally encoded into span-like structure, we can use a constituent-like parser for such a joint span tree.

\subsection{Semantic Representation}

Similar to the semantic representation of \cite{Li-aaai-19},
we use predicate-argument-relation tuples $\mathcal{Y} \in \mathcal{P} \times \mathcal{A} \times \mathcal{R}$, where $\mathcal{P} = \{w_1, w_2, ..., w_n\}$ is the set of all possible predicate tokens, $\mathcal{A} = \{(w_i,\dots,w_j) | 1 \leq i \leq j \leq n \}$ includes all the candidate argument spans and dependencies, and $\mathcal{R}$ is the set of the semantic roles and employ a null label $\epsilon$ to indicate no relation between predicate-argument pair candidate.
The difference from that of \cite{Li-aaai-19} is that in our model, we predict the span and dependency arguments at the same time which needs to distinguish the single word span arguments and dependency arguments.
Thus, we represent all the span arguments $\mathcal{A} = \{(w_i,\dots,w_j) | 1 \leq i \leq j \leq n \}$ as span S$(i-1, j)$ and all the dependency arguments $\mathcal{A} = \{(w_i) | 1 \leq i \leq n \}$ as span S$(i, i)$. We set a special start token at the beginning of sentence.

% In addition, as shown in Figure 2, we apply the SRL constraints that the label of span arguments is same as all the dependency arguments inside the span and all the arguments of a predicate cannot overlap. 
% To enforce the SRL constraints, we propose a single decoder with a dynamic programing to generate both dependency and span SRL at the same time.
% Since dependency-style dataset (CoNLL-2009) \cite{hajic-etal-2009-conll} contains nominal predicate while span-style dataset (CoNLL-2005) \cite{carreras-marquez-2005-introduction} has not. We add all nominal predicate and its dependency arguments as single span into CoNLL-2005 dataset to meet the SRL constraints. And assign a special token $\epsilo$ behind the label to represent the new added span-style arguments.

\section{Our Model}

\subsection{Overview}

As shown in Figure \ref{fig3}, our model includes four modules: token representation, self-attention encoder, scorer module, and two decoders.
Using an encoder-decoder backbone, we apply self-attention encoder \cite{Vaswani17} that is modified by position partition \cite{Kitaev-2018-SelfAttentive}.
We take multi-task learning (MTL) approach  sharing the parameters of token representation and self-attention encoder.
Since we convert two syntactic representations as joint span structure and apply uniform semantic representation, we only need two decoders, one for syntactic tree based on joint span syntactic parsing algorithm \cite{zhou-zhao-2019-head},
%CKY-style algorithm \cite{Cocke1970Programming, Younger1975Recognition, Kasami1965AN}, 
another for uniform SRL.

\subsection{Token Representation}

In our model, token representation $x_i$ is composed of characters, words, and part-of-speech (POS) representation.
For character-level representation, we use CharLSTM \cite{ling-etal-2015-finding}.
For word-level representation, we concatenate randomly initialized and pre-trained word embeddings.
We concatenate character representation and word representation as our token representation $x_i$=[$x_{char}$;$x_{word}$;$x_{POS}$]. 

In addition, we also augment our model with 
% ELMo \cite{PetersN18-1202},  
BERT \cite{Jacobbert} or XLNet \cite{XLNet-Zhilin-2019} as the sole token representation to compare with other pre-training models.
Since BERT and XLNet are based on sub-word, we only take the last sub-word vector of the word in the last layer of BERT or XLNet as our sole token representation $x_i$.

\label{Token Representation}

\subsection{Self-Attention Encoder}

The encoder in our model is adapted from \cite{Vaswani17} and factor explicit content and position information in the self-attention process. The input matrices $X = [x_1, x_2, \dots , x_n ]$ in which $x_i$ is concatenated with position embedding are transformed by a self-attention encoder. We factor the model between content and position information both in self-attention sub-layer and feed-forward network, whose setting details follow \cite{Kitaev-2018-SelfAttentive}. 
\subsection{Scorer Module}

Since span and dependency SRL share uniform representation, we only need three types of scores: syntactic constituent span, syntactic dependency  head, and semantic role scores.

We first introduce the span representation $s_{ij}$ for both constituent span and semantic role scores.
We define the left end-point vector as concatenation of the adjacent token $\overleftarrow{pl_i} = [\overleftarrow{y_i};\overleftarrow{y_{i+1}}]$, which $\overleftarrow{y_i}$ is constructed by splitting in half the outputs from the self-attention encoder.
Similarly, the right end-point vector is $\overrightarrow{pr_i} = [\overrightarrow{y_{i+1}};\overrightarrow{y_{i}}]$.
Then, the span representation $s_{ij}$ is the differences of the left and right end-point vectors $s_{ij} = [\overrightarrow{pr_j}-\overleftarrow{pl_i}]$
\footnote{Since we use the same end-point span $s_{ij} = [\overrightarrow{pr_j}-\overleftarrow{pl_i}]$ to represent the dependency arguments for our uniform SRL, we distinguish the left and right end-point vector ($\overleftarrow{pl_i}$ and $\overrightarrow{pr_i}$) to avoid having the zero vector as a span representation $s_{ij}$.}.

\noindent \textbf{Constituent Span Score} \quad
We follow the constituent syntactic parsing \cite{zhou-zhao-2019-head,Kitaev-2018-SelfAttentive,Gaddy} to train constituent span scorer.
We apply one-layer feedforward networks to generate span scores vector, taking span vector $s_{ij}$ as input:
$$
S(i, j) \&= W_2 g(LN( W_1s_{ij}+ b_1)) + b_2,
$$
\noindent where $LN$ denotes Layer Normalization, $g$ is the Rectified Linear Unit nonlinearity.
The individual score of category $\ell$ is denoted by
$$
S_{categ}(i, j, \ell) = [S(i, j)]_\ell, 
$$
\noindent where $[]_{\ell}$ indicates the value of corresponding the l-th element $\ell$ of the score vector.
The score $s(T)$ of the constituent parse tree $T$ is obtained by adding all scores of span ($i$, $j$) with category $\ell$: 
$$s(T) = \sum_{(i,j,\ell)\in T} S_{categ}(i, j, \ell).$$
The goal of constituent syntactic parsing is to find the tree with the highest score:
$\hat{T} = \arg\max_T s(T).$
We use CKY-style algorithm \cite{Gaddy} to obtain the tree $\hat{T}$ in $O(n^3)$ time complexity.
This structured prediction problem is handled with satisfying the margin constraint:
$$
s(T^*) \ge s(T) + \Delta (T,T^*),
$$
\noindent where $T^*$ denotes correct parse tree, and $\Delta$ is the Hamming loss on category spans with a slight modification during the dynamic programming search. The objective function is the hinge loss,
$$J_1(\theta) = \max ( 0,\max_T[s(T) + \Delta (T,T^*)]-s(T^*) ).$$
\noindent \textbf{Dependency Head Score} \quad
We predict a the possible heads and use the biaffine attention mechanism \cite{Dozat2017Deep} to calculate the score as follow:
$$
\alpha_{ij} = h_i^TWg_j + U^Th_i + V^T g_j + b,
$$
\noindent where $\alpha_{ij}$ indicates the child-parent score, $W$ denotes the weight matrix of the bi-linear term, $U$ and $V$ are the weight vectors of the linear term, and $b$ is the bias item, $h_i$ and $g_i$ are calculated by a distinct one-layer perceptron network.

We minimize the negative log-likelihood of the golden dependency tree $Y$, which is implemented as a cross-entropy loss:
$$
J_2(\theta) = -\left(logP_{\theta}(h_i|x_i) +logP_{\theta}(l_i|x_i,h_i)\right),
$$
\noindent where $P_{\theta}(h_i|x_i)$ is the probability of correct parent node $h_i$ for $x_i$, and $P_{\theta}(l_i|x_i,h_i)$ is the probability of the correct dependency label $l_i$ for the child-parent pair $(x_i,h_i)$.

\noindent \textbf{Semantic Role Score} \quad
To distinguish the currently considered predicate from its candidate arguments in the context, we employ one-layer perceptron to contextualized representation for argument $a_{ij}$\footnote{When $i$=$j$, it means a uniform representation of dependency semantic role.} candidates:
$$a_{ij} = g(W_3s_{ij}+ b_1),$$
\noindent where $g$ is the Rectified Linear Unit nonlinearity and $s_{ij}$ denotes span representation.

And predicate candidates $p_k$ is simply represented by the outputs from the self-attention encoder:
$p_k = y_k$.

For semantic role, different from \cite{Li-aaai-19}, we simply adopt concatenation of predicates and arguments representations, and one-layer feedforward networks to generate semantic role score:
$$
\Phi_{r}(p, a) \&= W_5 g(LN( W_4[p_k;a_{ij}]+ b_4)) + b_5,
$$
and the individual score of semantic role label $r$ is denoted by:
$$
\Phi_{r}(p, a, r) = [\Phi_{r}(p, a)]_r.
$$
Since the total of predicate-argument pairs are $O(n^3)$, which is computationally impractical.
We apply candidates pruning method in \cite{Li-aaai-19,he-etal-2018-jointly}.
First of all, we train separate scorers ($\phi_p$ and $\phi_a$) for predicates and arguments by two one-layer feedforward networks.
Then, the predicate and
argument candidates are ranked according to their predicted
score ($\phi_p$ and $\phi_a$), and we select the top $n_p$ and $n_a$ predicate and
argument candidates, respectively:
$$
n_p = \min(\lambda_pn, m_p),
n_a = \min(\lambda_an, m_a),
$$
\noindent where $\lambda_p$ and $\lambda_a$ are pruning rate, and $m_p$ and $m_a$ are maximal numbers of candidates.

Finally, the semantic role scorer is trained to optimize the probability $\textit{P}_\theta(\hat{y}|s)$ of the predicate-argument-relation tuples $\hat{y}_{(p,a,r)} \in \mathcal{Y}$ given the sentence $s$, which can be factorized as:
\begin{align}
J_3(\theta) \&= \sum_{p \in \mathcal{P},a \in \mathcal{A},r \in \mathcal{R}} -log\textit{P}_\theta(y_{(p,a,r)}|s)\nonumber\\
\&= \sum_{p \in \mathcal{P},a \in \mathcal{A},r \in \mathcal{R}} -log \frac{\exp{\phi(p, a,r)}}{\sum_{\hat{r} \in \mathcal{R}}\exp{\phi(p, a,\hat{r})}}\nonumber
\end{align}
where $\theta$ represents the model parameters, and $\phi(p, a, r) = \phi_p + \phi_a + \Phi_{r}(p, a, r)$ is the score by the predicate-argument-relation tuple including predicate score $\phi_p$, argument score $\phi_a$ and semantic role label score $\Phi_{r}(p, a, r)$.
In addition, we fix the score of null label $\phi(p, a, \epsilon) = 0$.

At last, we train our scorer for simply minimizing the overall loss:
$$
J_{overall}(\theta) = J_1(\theta) + J_2(\theta) + J_3(\theta).
$$

\subsection{Decoder Module}

\begin{algorithm}[t!]
\small
  \caption{Joint span syntactic parsing algorithm}
  \label{alg1}
  \begin{algorithmic}
  \REQUIRE sentence leng $n$, span and dependency score $s(i,j,\ell)$, $d(r,h)$, $1\leq i\leq j \leq n, \forall r,h,\ell$
  \ENSURE maximum value $S_H(T)$ of tree $T$
  \STATE \textbf{Initialization:} 
  \STATE $s_{c}[i][j][h] = s_{i}[i][j][h] = 0,  \forall i,j,h $ 
  \FOR{$len=1$ to $n$}
    \FOR{$i=1$ to $n-len + 1$}
        \STATE $j = i + len - 1$
        \IF {$len=1$}
            \STATE $
            s_{c}[i][j][i] = s_{i}[i][j][i] = \max_{\ell} s(i,j,\ell)
            $
        \ELSE
        \FOR{$h=i$ to $j$}
            \STATE 
            $\begin{aligned}
                split_l = &\max_{i\leq r < h} \ \{\ \max_{r \leq k < h}\ \{\ s_{c}[i][k][r] + \\
                &s_{i}[k+1][j][h]\ \} + d(r,h)\ \}
            \end{aligned}$
            
            \STATE  $\begin{aligned}
            split_r =& \max_{h < r \leq j}\ \{\ \max_{h \leq k < r}\ \{\ s_{i}[i][k][h] + \\
            &s_{c}[k+1][j][r]\ \} + d(r,h)\ \}
            \end{aligned}$
            
            \STATE $\begin{aligned}
            s_{c}[i][j][h] = & \max \ \{\ split_l, split_r\ \} + \\
            &\max_{\ell \neq \varnothing} s(i,j,\ell)
            \end{aligned}$
            
            \STATE $\begin{aligned}
            s_{i}[i][j][h] = &\max \ \{\ split_l, split_r\ \} + \\
            & \max_{\ell} s(i,j,\ell)
            \end{aligned}$
        \ENDFOR
        \ENDIF
   \ENDFOR
   \ENDFOR
   \STATE $
   S_H(T) =  \max_{1 \le h \le n}\ \{\ s_{c}[1][n][h] + d(h,root)\ \}
   $
  \end{algorithmic}
\end{algorithm}

\noindent  \textbf{Decoder for Joint Span Syntax}\quad

As the joint span is defined in a recursive way, to score the root joint span has been equally scoring all spans and dependencies in syntactic tree.

During testing, we apply the $joint$ $span$ CKY-style algorithm \cite{zhou-zhao-2019-head}, as shown in Algorithm \ref{alg1} to explicitly find the globally highest score $S_H(T)$ of our joint span syntactic tree $T$\footnote{For further details, see \cite{zhou-zhao-2019-head} which has discussed the different between constituent syntactic parsing CKY-style algorithm, how to binarize the $joint$ $span$ tree and the time, space complexity.}.

% In order to binarize the constituent parse tree with head, we 
% introduce the complete span $s_c$ and the incomplete span $s_i$, which is similar to Eisner algorithm \cite{EisnerP96}.
% The complete span except a single word only can be assigned not empty category $\O$.
% While the incomplete span and complete span consisting of a single word can be assigned any category including empty category $\O$.
% After finding the largest score $S_H(T)$, we backtrack the chart with split point $k$ and sub-root $r$ to construct the joint span syntactic tree $T$. 

% Comparing with constituent syntactic parsing CKY-style algorithm \cite{SternP17}, the dependency score $d(r,h)$ in our algorithm affects the selection of best split point $k$.
% Since we need to find the best value of sub-head $r$ and split point $k$,
% the complexity of the algorithm is $O(n^5)$ time\footnote{The syntactic parsing speed of our joint span decoder is 158.7 sents/sec on PTB which is not much slower than other parsers and suggests that training and inference
% times are dominated by neural network computations.} and $O(n^3)$ space.
% Since every spans contain a head word, we define the state $s_{com}(i,j,h)$ and $s_{incom}(i,j,h)$ as globally highest score of complete and incomplete span $(i,j)$ with head $h$.
Also, to control the effect of combining span and dependency scores, we apply a weight $\lambda_H$\footnote{We also try to incorporate the head information in constituent syntactic training process, namely max-margin loss for both two scores, but it makes the training process become more complex and unstable. Thus we employ a parameter to balance two different scores in joint decoder which is easily implemented with better performance.}:
$$
s(i,j,\ell) = \lambda_H  S_{categ}(i,j,\ell),
d(i,j) = (1 - \lambda_H )  \alpha_{ij},
$$

\noindent where $\lambda_H$ in the range of 0 to 1.
In addition, we can merely generate constituent or dependency syntactic parsing tree by setting $\lambda_H$ to 1 or 0, respectively.
\label{Joint Span Decoder}

\noindent \textbf{Decoder for Uniform Semantic Role}\quad
%Following the definition of the semantic roles, there are some dependencies between them. Since the predicted predicate-argument-relation tuples from the model may violate the SRL constraints described in
Since we apply uniform span for both dependency and span semantic role, 
we use a single dynamic programming decoder to generate two semantic forms following the non-overlapping constraints: span semantic arguments for the same predicate do not overlap \cite{punyakanok-2008-importance}.
% Algorithm \ref{alg2} shows the 

\section{Experiments}

We evaluate our model on CoNLL-2009 shared
task \cite{hajic-etal-2009-conll} for dependency-style SRL, CoNLL-2005 shared task \cite{carreras-marquez-2005-introduction} for span-style SRL both using the Propbank convention \cite{palmer-etal-2005-proposition}, and English Penn Treebank (PTB) \cite{MarcusJ93-2004} for constituent syntactic parsing, Stanford basic dependencies (SD) representation \cite{Marieffe06generatingtyped} converted by the Stanford parser{\footnote{http://nlp.stanford.edu/software/lex-parser.html}} for dependency syntactic parsing.
We follow standard data splitting: semantic (SRL) and syntactic parsing take  section 2-21 of Wall Street Journal (WSJ) data as training
set, SRL takes section 24 as development set while syntactic parsing takes section 22 as development set, SRL takes section 23 of WSJ together with 3 sections from Brown corpus as test set while syntactic parsing only takes section 23.
POS tags are predicted using the Stanford tagger \cite{Toutanova:2003}.
In addition, we use two SRL setups:
end-to-end and pre-identified predicates.

For the predicate disambiguation task in
dependency SRL, we follow \cite{marcheggiani-titov-2017-encoding} and use the off-the-shelf disambiguator
from \cite{roth-lapata-2016-neural}. 
For constituent syntactic parsing, we use the standard evalb{\footnote{http://nlp.cs.nyu.edu/evalb/}} tool to evaluate the F1 score. For dependency syntactic parsing, following previous work \cite{Dozat2017Deep}, we report the results without punctuations of the labeled and unlabeled attachment
scores (LAS, UAS).

\subsection{Setup}

\textbf{Hyperparameters}\quad In our experiments, we use 100D GloVe \cite{PenningtonD14-1162} pre-trained embeddings. For the self-attention encoder, we set 12 self-attention layers and use the same other hyperparameters settings as \cite{Kitaev-2018-SelfAttentive}.
For semantic role scorer, we use 512-dimensional MLP layers and 256-dimensional feed-forward networks.
For candidates pruning, we set $\lambda_p$ = 0.4 and $\lambda_a$ = 0.6 for pruning predicates and arguments, $m_p$ = 30 and $m_a$ = 300 for max numbers of predicates and arguments respectively.
For constituent span scorer, we apply a hidden size of 250-dimensional feed-forward networks.
For dependency head scorer, we employ two 1024-dimensional MLP layers with the ReLU as the activation function for learning specific representation and a 1024-dimensional parameter matrix for biaffine attention. 

In addition, when augmenting our model with 
% ELMo,  
BERT and XLNet, we set 
% 4 layers of self-attention for ELMo and
2 layers of self-attention for BERT and XLNet.

\noindent \textbf{Training Details}\quad we use 0.33 dropout for biaffine attention and MLP layers. All models are trained for up to 150 epochs with batch size 150 on a single NVIDIA GeForce GTX 1080Ti GPU with Intel i7-7800X CPU. 
We use the same training settings as \cite{Kitaev-2018-SelfAttentive} and \cite{kitaev2018multilingual}.

% \subsection{Span Representation}

% As mentions in section 
% Different token representation combinations are evaluated in Table \ref{table1}. We find that CharLSTM performs a little better than CharCNNs. Moreover, POS tags on parsing performance show that predicted POS tags decreases parsing accuracy, especially without word information. 
% If POS tags are replaced by word embeddings, the performance increases.
% Finally, we apply word and CharLSTM as token representation setting for our full model{\footnote{We also evaluate POS tags on CTB which increases parsing accuracy, thus we employ the word, POS tags and CharLSTM as token representation setting for CTB.}}.

\begin{figure}[t!]
    \centering
    \includegraphics[width=3.2in]{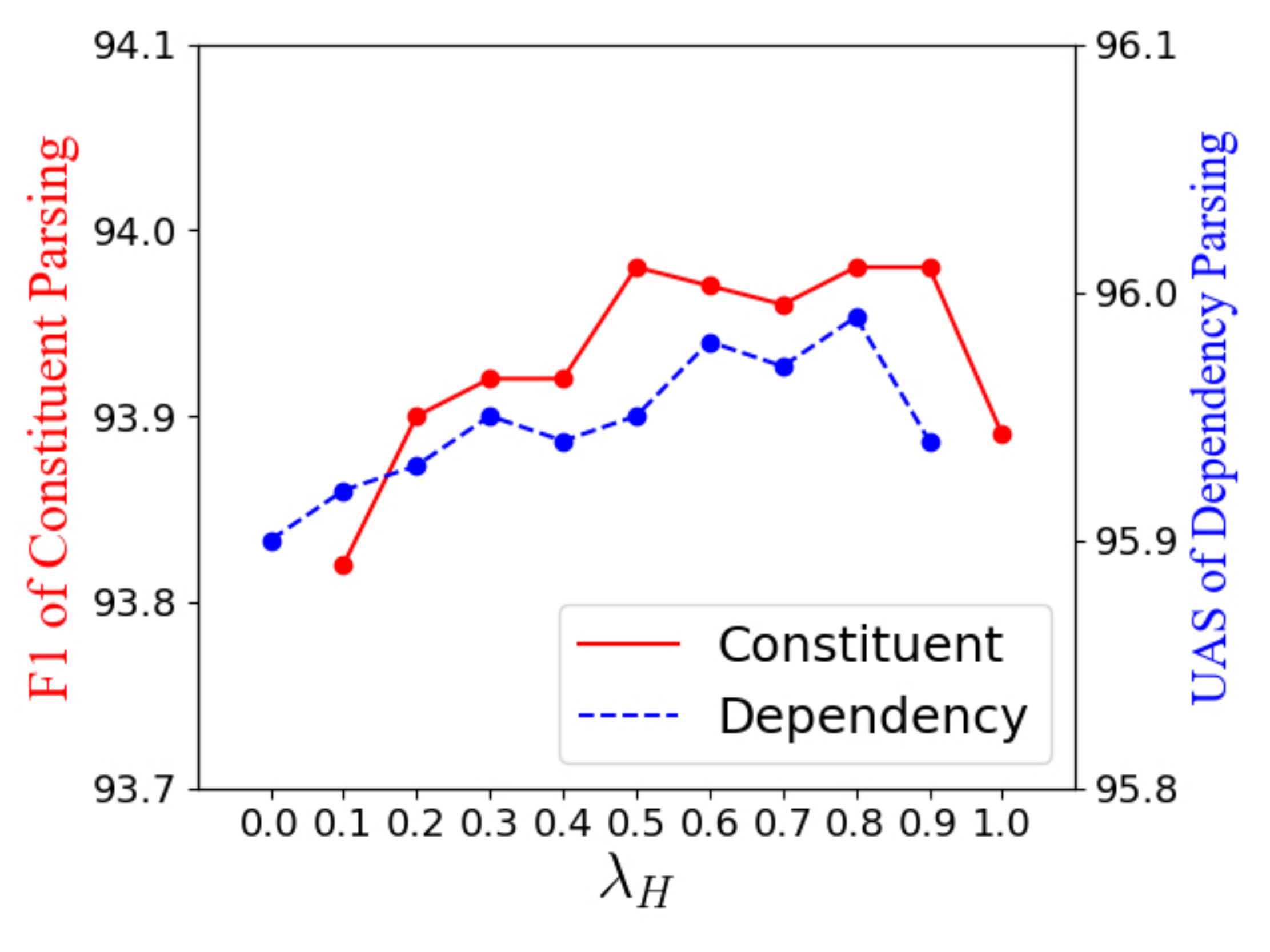}
    
    \caption{Syntactic parsing performance of different parameter $\lambda_H$ on PTB dev set.}
    \label{lambda}
\end{figure}

\begin{table}[t!]
    \begin{center}
    \small
    \resizebox{\linewidth}{!}{  
        \begin{tabular*}{\hsize}{@{}@{\extracolsep{\fill}}lccc@{}}
            \hline
            \bf Model                           &F1 &UAS &LAS\\
            \hline
            separate constituent &\multirow{2}{0.7cm}{93.98}&$-$&$-$ \\
            converted dependency &         &95.38 &94.06 \\
            separate dependency &$-$&95.80&94.40 \\
            
            \hline
            joint span $\lambda_H$ = 1.0   &93.89 &$-$  &$-$ \\
            joint span $\lambda_H$ = 0.0   &$-$  &95.90 &94.50 \\
%           joint span $\lambda_H$ = 0.5   &\multirow{2}{0.7cm}{\bf93.98} &\bf95.95 & 94.51 \\
%           converted dependency   &  &95.67 &\bf 94.53 \\
            joint span $\lambda_H$ = 0.8   &\multirow{2}{0.7cm}{\bf93.98} &\bf95.99 & 94.53 \\
            converted dependency   &  &95.70 &\bf 94.60 \\
            \hline
        \end{tabular*}}
    \end{center}
    \caption{\label{table1} PTB dev set performance of joint span syntactic parsing. The \textit{converted} means the corresponding dependency syntactic parsing results are from the corresponding constituent parse tree using head rules.}
\end{table}
\subsection{Joint Span Syntactic Parsing}

This subsection examines joint span syntactic parsing decoder \ref{Joint Span Decoder} with semantic parsing both of dependency and span. 
The weight parameter $\lambda_H$ plays an important role to balance the syntactic span and dependency scores.
When $\lambda_H$ is set to 0 or 1, the joint span parser works as the dependency-only parser or constituent-only parser respectively.
$\lambda_H$ set to between 0 to 1 indicates the general joint span syntactic parsing, providing both constituent and dependency structure prediction.
We set the $\lambda_H$ parameter from 0 to 1 increased by 0.1 step as shown in Figure \ref{lambda}. The best results are achieved when $\lambda_H$ is set to 0.8 which achieves the best performance of both syntactic parsing. 

In addition, we compare the joint span syntactic parsing decoder with a separate learning constituent syntactic parsing model which takes the same token representation, self-attention encoder and joint learning setting of semantic parsing on PTB dev set.
The constituent syntactic parsing results are also converted into dependency ones by PTB-SD for comparison.

Table \ref{table1} shows that joint span decoder benefit both of  constituent and dependency syntactic parsing.
Besides, the comparison also shows that the directly predicted dependencies from our model are better than those converted from the predicted constituent parse trees in UAS term.
% that even in dependency-only or constituent-only mode, our joint span parser still outperforms the separate constituent parser in terms of either constituent and dependency parsing performance.
% Besides, the comparison also shows that the directly predicted dependencies from our model are slightly better than those converted from the predicted constituent parse trees.

% \begin{table}[t!]
%   \begin{center}
%   \small
%   \resizebox{\linewidth}{!}{  
%       \begin{tabular*}{\hsize}{@{}@{\extracolsep{\fill}}lccc@{}}
%           \hline
%           \bf Model                           &F1 &UAS &LAS\\
%           \hline
%           separate constituent &\multirow{2}{0.7cm}{93.47}&$-$&$-$ \\
%           converted dependency &         &95.06 &93.81 \\
%           \hline
%           joint span $\lambda_H$ = 1.0   &93.67 &$-$  &$-$ \\
%           joint span $\lambda_H$ = 0.0   &$-$  &95.82 &94.43 \\
%           joint span $\lambda_H$ = 0.5   &\multirow{2}{0.7cm}{\bf93.78} &\bf95.92 &\bf94.49 \\
%           converted dependency   &  &95.69 &94.45 \\
%           \hline
%       \end{tabular*}}
%   \end{center}
%   \caption{\label{table3} PTB dev set performance of joint span syntactic parsing. The \textit{converted} means the corresponding dependency syntactic parsing results are from the corresponding constituent parse tree using head rules.}
% \end{table}

% \begin{figure}[t!]
%     \centering
%     \includegraphics[width=2.8in]{lambda.eps}
    
%     \caption{Balancing constituent and dependency of joint span HPSG parsing on English dev set.}
%     \label{fig4}
% \end{figure}

\begin{table}[t!]
    % \small
    \centering
    \resizebox{\linewidth}{!}{
    \begin{tabular}{lccccc}  
        \toprule  
        \multirow{2}{*}{System}
        &\multicolumn{1}{c}{SEM$_{span}$}
        &\multicolumn{1}{c}{SEM$_{dep}$}
        &\multicolumn{1}{c}{SYN$_{con}$}
        &\multicolumn{2}{c}{SYN$_{dep}$}
        \cr  
        \cmidrule(lr){2-2} 
        \cmidrule(lr){3-3} 
        \cmidrule(lr){4-4}
        \cmidrule(lr){5-6}
         &F$_1$&F$_1$&F$_1$&UAS&LAS\cr
    \midrule
    \textit{End-to-end} \cr
    SEM$_{span}$ &82.27 &$-$ &$-$ &$-$ &$-$\cr
    SEM$_{dep}$ &$-$ &84.90 &$-$ &$-$ &$-$\cr
    SEM$_{span,dep}$ &83.50 &84.92 &$-$ &$-$ &$-$\cr
    % Con &$-$&$-$&$-$&$-$&$-$&$-$&93.40 &93.60 &93.50 &- &-\cr
    % Dep &$-$&$-$&$-$&$-$&$-$&$-$&- &- &- &95.55 &94.15 \cr
    SEM$_{span,dep}$, SYN$_{con}$ &\bf83.81&\bf 84.95 &\bf93.98 &$-$ &$-$\cr
    SEM$_{span,dep}$, SYN$_{dep}$ &83.13&84.24 &$-$ &95.80 &94.40\cr
    % Con, Dep &$-$&$-$&$-$&$-$&$-$&$-$&93.52 &93.59 &93.56 &95.70 &94.38\cr
    SYN$_{con,dep}$ &$-$&$-$ &93.78 &95.92 &94.49\cr
    SEM$_{span,dep}$, SYN$_{con,dep}$ &83.12 &83.90 &\bf93.98 &\bf95.95 &\bf94.51\cr
    
    \midrule 
    \textit{Given predicate}    \cr
    SEM$_{span}$ &83.16 &$-$ &$-$ &$-$ &$-$\cr
    SEM$_{dep}$ &$-$ &88.23 &$-$ &$-$ &$-$\cr
    SEM$_{span,dep}$ & 84.74 &88.32  &$-$ &$-$ &$-$\cr
    SEM$_{span,dep}$, SYN$_{con}$ &84.46&\bf 88.40&93.78 &$-$ &$-$\cr
    SEM$_{span,dep}$, SYN$_{dep}$ &\bf 84.76 &87.58 &$-$ &95.94 &94.54\cr
    SEM$_{span,dep}$, SYN$_{con,dep}$ &84.43 &87.58 &\bf 94.07 &\bf 96.03 &\bf 94.65\cr
        \bottomrule  
    \end{tabular}
    }
    \caption{Joint learning analysis on CoNLL-2005, CoNLL-2009, and PTB dev sets.}\label{Joint Learning Analysis}
\end{table}

\subsection{Joint Learning Analysis} 

% are more effective than learning constituent or dependency parsing separately.
Table \ref{Joint Learning Analysis} compares the different joint setting of semantic (SRL) and syntactic parsing
to examine whether semantics and syntax can enjoy their joint learning.
In the end-to-end mode, we find that constituent syntactic parsing can boost both
styles of semantics while dependency syntactic parsing cannot.
Moreover, the results of the last two rows indicate that semantics can benefit syntax simply by optimizing the joint objectives.
While in the given predicate mode, both constituent and dependency syntactic parsing can enhance SRL.
In addition, joint learning of our uniform SRL performs better than separate learning of either dependency or span SRL in both modes.

Overall, joint semantic and constituent syntactic parsing achieve relatively better SRL results than the other settings.
Thus, the rest of the experiments are done with multi-task learning of semantics and constituent syntactic parsing (wo/dep).
Since semantics benefits both of two syntactic formalisms and two syntactic parsing can benefit each other, we also compare the results of joint learning with semantics and two syntactic parsing models (w/dep).

% In addition, we evaluate the contribution of different components by ablation study as shown in Table \ref{ablation study}.
% The results indicate that span and dependency SRL are both important for each other, which is similar to syntactic parsing. 
% Since SRL can benefit both of two syntactic formalisms and joint SRL and constituent parsing achieve best results, we choose two settings for comparation:
% multi-task learning of SRL, constituent parsing, with or without dependency parsing.

\subsection{Syntactic Parsing Results}

In the wo/dep setting, we convert constituent syntactic parsing results into dependency ones by PTB-SD for comparison and set $\lambda_H$ described in \ref{Joint Span Decoder} to 1 for generating constituent syntactic parsing only.

Compared to the existing state-of-the-art models without pre-training, our performance exceeds \cite{zhou-zhao-2019-head} nearly 0.2 in LAS of dependency and 0.3 F1 of constituent syntactic parsing which are considerable improvements on such strong baselines.
% our model and our model achieves new state-of-the-art on both constituent and dependency syntactic parsing without pre-training.
Compared with \cite{strubell-etal-2018-linguistically} shows that our joint model setting boosts both of syntactic parsing and SRL which are consistent with \cite{shi-etal-2016-exploiting} that syntactic parsing and SRL benefit relatively more from each other.

% In addition, we first report constituent parsing result of our models on
% the CoNLL-2005 Brown test set.

% The syntactic parsing results demonstrate that joint span HPSG decoder indeed impoves syntactic parsing performance by incorporating phrase and head information.
%are more effective than learning constituent or dependency parsing separately.

We augment our parser with 
% ELMo, 
a larger version of BERT and XLNet as the sole token representation to compare with other models.
Our single model in XLNet setting achieving 96.18 F1 score of constituent syntactic parsing, 97.23\% UAS and 95.65\% LAS of dependency syntactic parsing.

\begin{table}[t!]
    \begin{center}
    \small
    \resizebox{\linewidth}{!}{  
        \begin{tabular}{lcc}
            \hline
%           \multirow{2}{*}{} & \multicolumn{2}{c}{WSJ} \\
%           \cline{2-3}
            \multirow{2}{*}&UAS &LAS \\
            \hline
            %\cite{MaI17-1007} &94.88 &92.98\\
            \citet{Dozat2017Deep} &95.74 &94.08 \\
            \citet{Ma2018Stack}  &95.87 &94.19 \\
            \citet{strubell-etal-2018-linguistically}&94.92 &91.87  \\
            \citet{Daniel-2019-naacl-left}&96.04 &94.43 \\
            \citet{zhou-zhao-2019-head} &96.09 &94.68 \\
            \hline
            \bf Ours converted  (wo/dep) &95.20 &93.90  \\
%           \bf Ours (w/dep)  &\bf96.11 &\bf94.82 \\
            \bf Ours (w/dep)  &\bf96.15 &\bf94.85 \\
            \hline
            \bf Pre-training \\
%           \cite{WangD18-1311}(ELMo) &96.35 &95.25  \\
            \citet{strubell-etal-2018-linguistically} &96.48 &94.40  \\
            \citet{ji-etal-2019-graph} &95.97 &94.31 \\
            \citet{zhou-zhao-2019-head} &97.00 &95.43\\
%           \cite{zhou-zhao-2019-head} &97.20 &95.72\\
            \hline
%           \bf Our converted  (wo/dep) + ELMo  &96.21 &95.02  \\
%           \bf Our (w/dep) + ELMo  &96.72 &95.00  \\
            \bf Ours converted  (wo/dep) + BERT  &96.77 & 95.72 \\
            \bf Ours (w/dep) + BERT  & 96.90 &95.32  \\
            \bf Ours converted  (wo/dep) + XLNet  &97.21 &\bf 96.25 \\
            \bf Ours (w/dep) + XLNet  & \bf 97.23 &95.65  \\
            \hline
        \end{tabular}}
    \end{center}
    \caption{\label{syndep} Dependency syntactic parsing on WSJ test set.}
\end{table}

% \begin{table}[t!]
%   \begin{center}
%   \small
%   \resizebox{\linewidth}{!}{  
%       \begin{tabular}{lcccc}
%           \hline
%           \multirow{2}{*}{} & \multicolumn{2}{c}{WSJ} & \multicolumn{2}{c}{Brown} \\
%           \cline{2-3}
%           \cline{4-5}
%           \multirow{2}{*}&UAS &LAS &UAS &LAS\\
%           \hline
%           \citet{MaI17-1007} &94.88 &92.98 &$-$ &$-$\\
%           \citet{Dozat2017Deep} &95.74 &94.08 &$-$ &$-$\\
%           \citet{Ma2018Stack}  &95.87 &94.19 &$-$ &$-$\\
%           \citet{Daniel-2019-naacl-left}&96.04 &94.43 &$-$ &$-$ \\
%           \citet{strubell-etal-2018-linguistically}&94.92 &91.87 &90.31 &85.82 \\
%           \hline
%           \bf Ours (wo/dep) &- &- &- &- \\
%           \bf Ours (w/dep)  &\bf96.11 &\bf94.82 &- &- \\
%           \hline
%           \bf Pre-training \\
%           \citet{WangD18-1311}(ELMo) &96.35 &95.25 &\_ &\_ \\
%           \citet{strubell-etal-2018-linguistically}(ELMo) &96.48 &94.40 &92.56 &88.52 \\
%           \hline
%           \bf Our (wo/dep) + ELMo  &- &- &- &- \\
%           \bf Our (w/dep) + ELMo  &96.72 &95.00 &- &- \\
%           \bf Ours (wo/dep) + BERT  &- &- &- &- \\
%           \bf Ours (w/dep) + BERT  &96.9 &- &- &- \\
%           \hline
%       \end{tabular}}
%   \end{center}
%   \caption{\label{table4} Dependency parsing on WSJ and Brown test set.}
% \end{table}

\begin{table}[t!]
    \centering
    \small
        \resizebox{\linewidth}{!}{
        \begin{tabular*}{\hsize}{@{}@{\extracolsep{\fill}}lccc@{}}
            \hline
                           &LR &LP &F1\\
            \hline 
            % \cite{SternP17}&93.2 &90.3 &91.8\\
            \citet{Gaddy} &91.76 &92.41 &92.08\\
            \citet{SternD17b} &92.57 &92.56 &92.56\\
            \citet{Kitaev-2018-SelfAttentive}  &93.20  &93.90 &93.55\\
            \citet{zhou-zhao-2019-head} &93.64 &93.92 &93.78 \\
            \hline
            \bf Ours (wo/dep)     &93.56  &94.01 &93.79 \\
%           \bf Ours (w/dep)    &93.89  &94.10 &\bf94.00 \\
            \bf Ours (w/dep)    &93.94  &94.20 &\bf94.07 \\
            \hline
            \bf Pre-training \\
            \citet{Kitaev-2018-SelfAttentive} &94.85 &95.40 &95.13\\
            \citet{kitaev2018multilingual} &95.46 &95.73 &95.59\\
            \citet{zhou-zhao-2019-head} &95.70 &95.98 &95.84 \\
            % \cite{zhou-zhao-2019-head} &96.21 &96.46 96.33\\

            \hline
%           \bf Ours (wo/dep) + ELMo    &94.73  &95.13 &94.93 \\
%           \bf Ours (w/dep) + ELMo    &95.07  &95.40 &95.23 \\
            \bf Ours (wo/dep) + BERT   &95.27  &95.76 &95.51 \\
            \bf Ours (w/dep) + BERT   &95.39  &95.64 &95.52 \\
            \bf Ours (wo/dep) + XLNet   &96.01  &96.36 &\bf96.18 \\
            \bf Ours (w/dep) + XLNet   &96.10  &96.26 &\bf96.18 \\
            
            \hline
        \end{tabular*}}
    \caption{\label{synconst} Constituent syntactic parsing on WSJ test set}
\end{table}

\begin{table}[t!]
    \centering
    \resizebox{\linewidth}{!}{
    \begin{tabular}{lcccccc}  
        \toprule  
        \multirow{2}{*}{System}
        &\multicolumn{3}{c}{WSJ}
        &\multicolumn{3}{c}{Brown}
        \cr  
        \cmidrule(lr){2-4} 
        \cmidrule(lr){5-7} 
         &P&R&F$_1$&P&R&F$_1$\cr
    \midrule
    \textit{End-to-end} \cr
%   \cite{he-etal-2017-deep} &80.2 &82.3 &81.2 &67.6 &69.6 &68.5\cr
    \citet{he-etal-2018-jointly} &81.2 &83.9 &82.5 &69.7 &71.9 &70.8\cr
    \citet{Li-aaai-19} &- &- &83.0 &- &- &-\cr
    \citet{strubell-etal-2018-linguistically}
    &84.07 &83.16 &83.61 &73.32 &70.56 &71.91\cr
    \citet{strubell-etal-2018-linguistically}*
    &85.53 &84.45 &\bf 84.99 &75.8 &73.54 &\bf 74.66\cr
    \midrule
    \bf Ours (wo/dep) &83.65&85.48&84.56&72.02&73.08&72.55\cr
    \bf Ours (w/dep) &83.54&85.30&84.41&71.84&72.07&71.95\cr
    \midrule
    \bf + Pre-training \cr
    \citet{he-etal-2018-jointly} &84.8 &87.2 &86.0 &73.9 &78.4 &76.1\cr
    \citet{Li-aaai-19} &85.2 &87.5 &86.3 &74.7 &78.1 &76.4\cr
    \citet{strubell-etal-2018-linguistically}
    &86.69 &86.42 &86.55 &78.95 &77.17 &78.05\cr
    \citet{strubell-etal-2018-linguistically}*
    &87.13 &86.67 &86.90 &79.02 &77.49 &78.25\cr
   \midrule
%   \bf Ours (wo/dep) + ELMo &85.30&87.70&86.48&76.07&78.27&77.15\cr
%     \bf Ours (w/dep) + ELMo &85.33&87.70&86.50&75.95&78.30&77.11\cr
    \bf Ours (wo/dep) + BERT
    &86.77&88.49&87.62&79.06&81.67&80.34\cr
    \bf Ours (w/dep) + BERT &86.46&88.23& 87.34&77.26&80.20&78.70\cr
    \bf Ours (wo/dep) + XLNet
    &87.65&89.66&\bf88.64&80.77&83.92&\bf82.31\cr
    \bf Ours (w/dep) + XLNet 
    &87.48&89.51&88.48&80.46&84.15&82.26\cr
    \midrule 
    \midrule
    \textit{Given predicate}    \cr
%   \cite{he-etal-2017-deep} &83.1 &83.0 &83.1 &72.9 &71.4 &72.1\cr
    \citet{Tan-Deep-Semantic}&84.5 &85.2 &84.8 &73.5 &74.6 &74.1 \cr
    \citet{he-etal-2018-jointly}  &- &- &83.9 &- &- &73.7\cr
     \citet{ouchi-etal-2018-span}&84.7 &82.3 &83.5 &76.0 &70.4 &73.1\cr
    \citet{strubell-etal-2018-linguistically} 
    &84.72 &84.57 &84.64 &74.77 &74.32 &74.55\cr
    \citet{strubell-etal-2018-linguistically}* 
    &86.02 &86.05 &\bf 86.04 &76.65 &76.44 &\bf 76.54\cr
    \midrule
    \bf Ours (wo/dep) &85.93&85.76&85.84&76.92&74.55&75.72\cr
    
    \bf Ours (w/dep) &85.61&85.39&85.50&73.9&73.22&73.56\cr
    \midrule
    \bf + Pre-training \cr
    \citet{he-etal-2018-jointly}&- &- &87.4 &- &- &80.4\cr
    \citet{ouchi-etal-2018-span}&88.2 &87.0 &87.6 &79.9 &77.5 &78.7\cr
    \citet{Li-aaai-19}
    &87.9 &87.5 &87.7 &80.6 &80.4 &80.5\cr
    \midrule
    % \bf Ours (wo/dep) + ELMo &87.76&88.29&88.02&79.59&78.64&79.11\cr
    % \bf Ours (w/dep) + ELMo &87.75&87.91&87.82&80.81&79.51&80.15\cr
    \bf Ours (wo/dep) + BERT &89.04&88.79&88.91&81.89&80.98&81.43\cr
    \bf Ours (w/dep) + BERT &88.94&88.53& 88.73&81.66&80.80&
    81.23\cr
    \bf Ours (wo/dep) + XLNet
    &89.89&89.74&\bf89.81&85.35&84.57&\bf84.96\cr
    \bf Ours (w/dep) + XLNet 
    &89.62&89.82&89.72&85.08&84.84&\bf84.96\cr
    \bottomrule  
    \end{tabular}
    }
    \caption{Span SRL results on CoNLL-2005 test sets. * represents injecting state-of-the-art predicted parses.}\label{Span SRL results}
\end{table}

\begin{table}[t!]
    \small
    \centering
    \resizebox{\linewidth}{!}{
    \begin{tabular}{lcccccc}  
        \toprule  
        \multirow{2}{*}{System}
        &\multicolumn{3}{c}{WSJ}
        &\multicolumn{3}{c}{Brown}
        \cr  
        \cmidrule(lr){2-4} 
        \cmidrule(lr){5-7} 
         &P&R&F$_1$&P&R&F$_1$\cr
    \midrule
    \textit{End-to-end} \cr
%   \cite{zhao-2014-Integrative} &$-$ &$-$ &82.5 &$-$ &$-$ &$-$\cr
    \citet{Li-aaai-19} &- &- &85.1 &- &- &-\cr
    \midrule
    \bf Ours (wo/dep) &84.24&87.55&\bf 85.86&76.46&78.52&\bf 77.47\cr
    \bf Ours (w/dep) &83.73&86.94& 85.30&76.21&77.89& 77.04\cr
    \midrule
    \bf + Pre-training \cr
    \citet{he-etal-2018-syntax} &83.9 &82.7 &83.3 &- &- &-\cr
    \citet{cai-etal-2018-full}&84.7 &85.2 &85.0 &- &- &72.5\cr
    \citet{Li-aaai-19} &84.5 &86.1 &85.3 &74.6 &73.8 &74.2\cr
    \midrule
    % \bf Ours(wo/dep) + ELMo &85.21&88.17&86.66&78.62&80.76&79.68\cr
    % \bf Ours (w/dep) + ELMo &84.85&88.21& 86.50&78.43&80.52&79.46\cr
    \bf Ours (wo/dep) + BERT &87.40&88.96&88.17&80.32&82.89&81.58\cr
    \bf Ours (w/dep) + BERT &86.77&89.14& 87.94&79.71&82.40& 81.03\cr
    \bf Ours (wo/dep) + XLNet
    &86.58&90.40&\bf88.44&80.96&85.31&\bf83.08\cr
    \bf Ours (w/dep) + XLNet 
    &86.35&90.16&88.21&80.90&85.38&\bf83.08\cr
    \midrule
    \midrule 
    \textit{Given predicate}    \cr
%   \citet{marcheggiani-etal-2017-simple} &88.7 &86.8 &87.7 &79.4 &76.2 &77.7\cr
%   \citet{marcheggiani-titov-2017-encoding} &89.1 &86.8 &88.0 &78.5 &75.9 &77.2\cr
    \cite{kasai-etal-2019-syntax} &89.0 &88.2 & 88.6 &78.0 &77.2 &77.6\cr 
    \midrule
    \bf Ours (wo/dep) &88.73&89.83&\bf 89.28&82.46&83.20&\bf 82.82\cr
    \bf Ours (w/dep) &88.02&89.03&88.52&80.98&82.10&81.54\cr
    \midrule
    \bf + Pre-training \cr
    \citet{he-etal-2018-syntax}&89.7 &89.3 &89.5 &81.9 &76.9 &79.3\cr
    \citet{cai-etal-2018-full}&89.9 &89.2 &89.6 &79.8 &78.3 &79.0\cr
    \citet{Li-aaai-19} &89.6 &91.2 &90.4 &81.7 &81.4 &81.5\cr
    % \cite{Li-aaai-19} &90.0 &90.0 &90.0 &81.7 &81.4 &81.5\cr 
    \citet{kasai-etal-2019-syntax} &90.3 &90.0 &90.2 &81.0 &80.5 &80.8\cr 
    \citet{lyu-etal-2019-semantic} &- &- &90.99 &- &- &82.18 \cr
    \citet{chen-etal-2019-capturing}
    &90.74 &91.38 &91.06 &82.66 &82.78 &82.72 \cr 
    \citet{cai-lapata-2019-semi} &91.7 &90.8 &91.2  &83.2 &81.9 &82.5\cr 
    \midrule
    % \bf Ours (wo/dep) + ELMo &89.71&90.90&90.30&83.94&85.04&84.49\cr
    % \bf Ours (w/dep) + ELMo &89.38&90.26&89.82&83.96&84.80&84.38\cr
    \bf Ours (wo/dep) + BERT
    &91.21&91.19&91.20&85.65&86.09&85.87\cr
    \bf Ours (w/dep) + BERT &91.14&91.03&91.09&85.18&85.41& 85.29\cr
    \bf Ours (wo/dep) + XLNet
    &91.16&91.60&\bf91.38&87.04&87.54&\bf87.29\cr
    \bf Ours (w/dep) + XLNet 
    &90.80&91.74&91.27&86.43&87.25&86.84\cr
    \bottomrule  
    \end{tabular}
    }
    \caption{Dependency SRL results on CoNLL-2009 Propbank test sets.}\label{Dependency SRL results}
\end{table}
\subsection{Semantic Parsing Results}

We present all results using the official evaluation script from the CoNLL-2005 and CoNLL-2009 shared tasks, and compare our model with previous state-of-the-art models in Table \ref{Span SRL results}, \ref{Dependency SRL results}.
The upper part of the tables presents results from end-to-end mode while the lower part shows the results of given predicate mode to compare to more previous works with pre-identified predicates.
In given predicate mode, we simply replace predicate candidates with the gold predicates without other modification on the input or encoder.

\noindent \textbf{Span SRL Results}\quad Table \ref{Span SRL results} shows results on CoNLL-2005 in-domain (WSJ) and out-domain (Brown) test sets. 
It is worth noting that \cite{strubell-etal-2018-linguistically} injects state-of-the-art predicted parses in terms of setting of \cite{Dozat2017Deep} at test time and aims to use syntactic information to help SRL.
While our model not only excludes other auxiliary information during test time but also benefits both syntax and semantics.
We obtain comparable results with the
state-of-the-art method \cite{strubell-etal-2018-linguistically} and outperform all recent models without additional information in test time.
After incorporating with pre-training contextual representations, %ELMo or BERT, 
our model achieves new state-of-the-art both of end-to-end and given predicate mode and both of in-domain and out-domain. 

% Our model outperforms the previous models with absolute improvements in F$_1$-score of 0.3\% on CoNLL-2005 benchmark. Besides, our single model performs even much better than all previous ensemble systems.
% %On all datasets, our model is able to predict over 40\% of the sentences completely correctly

%\paragraph{Dependency SRL}
\noindent \textbf{Dependency SRL Results}\quad Table \ref{Dependency SRL results} presents the results on CoNLL-2009. We obtain new state-of-the-art both of end-to-end and given predicate mode and both of in-domain and out-domain text. 
These results demonstrate that our improved uniform SRL representation can be adapted to perform dependency SRL and achieves impressive performance gains.

\section{Related Work}

In the early work of SRL, most of the researchers focus on feature engineering based on training corpus.
The traditional approaches to SRL focused on developing
rich sets of linguistic features templates and then employ linear classifiers such as SVM \cite{zhao-etal-2009-multilingual-dependency}.
With the impressive success of deep neural networks
in various NLP tasks \cite{luo-zhao-2020-bipartite,Li2020Data-dependent, he-etal-2019-syntax, luo2019named,zhang-etal-2018-exploring,li-etal-2018-seq2seq, zhang-etal-2018-modeling,luo2019hierarchical,zhang-etal-2019-open,li-etal-2019-gan,zhao-kit-2008-parsing,zhao-etal-2009-semantic,Zhao_2013}, considerable attention has been paid to syntactic features \cite{strubell-etal-2018-linguistically,kasai-etal-2019-syntax,he-etal-2018-syntax}.

\cite{lewis-etal-2015-joint,strubell-etal-2018-linguistically,kasai-etal-2019-syntax,he-etal-2018-syntax,li-etal-2018-unified}
modeled syntactic
parsing and SRL jointly, \cite{lewis-etal-2015-joint}
jointly modeled SRL and CCG parsing, and \cite{kasai-etal-2019-syntax} combined the supertags extracted from dependency parses with SRL .

There are a few studies on joint learning of syntactic and semantic parsing which only focus on dependency structure \cite{swayamdipta-etal-2016-greedy,henderson-etal-2013-multilingual,shi-etal-2016-exploiting}.
Such as \cite{henderson-etal-2013-multilingual} based on dependency structure only focus on shared representation without explicitly analyzing whether syntactic and semantic parsing can benefit each other.
The ablation studies results show joint learning can benefit semantic parsing while the single syntactic parsing model was insignificantly worse (0.2\%) than the joint model.
\cite{shi-etal-2016-exploiting} only made a brief attempt on Chinese Semantic Treebank %\cite{qiu-2016} 
to show the mutual benefits between dependency syntax and semantic roles.
% The ablation studies results show joint learning can benefit semantic parsing while syntactic parsing performance are the same.
% In fact, this paper find that the single syntactic parsing model was non-significantly worse (0.2\%) than the joint model.
Instead, our work focuses on whether syntactic and semantic parsing can benefit each other both on span and dependency in a more general way.

Besides, both span and dependency are effective formal representations for both semantics and syntax.
On one hand, researchers are interested in two forms of SRL models that may benefit from each other rather than their separated development, which has been roughly discussed in \cite{johansson-nugues-2008-dependency}.
\cite{he-etal-2018-jointly} is the first to apply span-graph structure based on contextualized span representations to span SRL and \cite{Li-aaai-19} built on these span representations achieves state-of-art results on both span and dependency SRL using the same model but training individually.
On the other hand, researchers have discussed how to encode lexical dependencies in phrase structures, like lexicalized tree adjoining grammar (LTAG) \cite{SCHABESC88-2121}
% , Combinatory Categorial Grammar (CCG) \cite{steedman2000syntactic} 
and head-driven phrase structure grammar (HPSG) \cite{pollard1994head}.
% Thus, it is a natural idea to study the relationship between constituent and dependency structures, and the joint learning of constituent and dependency syntactic parsing \cite{KleinP04,CharniakP05-1022,FarkasW11-2924,GreenW12-0503,RenCombine2013,XuP14-1021,YoshikawaP17-1026}.

% To further exploit both strengths of the two representation forms and relationship of semantic and syntax, in this work, we propose a multi-task learning model that formulizes uniform SRL and constituent and dependency structures as simplified HPSG.

% Recently, constituent and dependency parsing have been well developed with neural network. These models attain state-of-the-art results for dependency parsing \cite{ChenD14, Dozat2017Deep, Ma2018Stack} and constituent parsing \cite{Dyer-N16-1024, Cross, Kitaev-2018-SelfAttentive}. 

\section{Conclusions}

This paper presents the first joint learning model which is evaluated on four tasks: span and dependency SRL, constituent and dependency syntactic parsing.
We exploit the relationship between semantics and syntax and conclude that not only syntax can help semantics but also semantics can improve syntax performance.
Besides, we propose two structure representations, uniform SRL and joint span of syntactic structure, to combine the span and dependency forms.
From experiments on these four parsing tasks, our single model achieves state-of-the-art or competitive results.

\bibliography{anthology,emnlp2020}

\begin{thebibliography}{59}
\expandafter\ifx\csname natexlab\endcsname\relax\def\natexlab#1{#1}\fi

\bibitem[{Cai et~al.(2018)Cai, He, Li, and Zhao}]{cai-etal-2018-full}
Jiaxun Cai, Shexia He, Zuchao Li, and Hai Zhao. 2018.
\newblock \href {https://www.aclweb.org/anthology/C18-1233} {A full end-to-end
  semantic role labeler, syntactic-agnostic over syntactic-aware?}
\newblock In \emph{Proceedings of the 27th International Conference on
  Computational Linguistics}, pages 2753--2765, Santa Fe, New Mexico, USA.
  Association for Computational Linguistics.

\bibitem[{Cai and Lapata(2019)}]{cai-lapata-2019-semi}
Rui Cai and Mirella Lapata. 2019.
\newblock \href {https://www.aclweb.org/anthology/D19-1094} {{Semi-Supervised
  Semantic Role Labeling with Cross-View Training}}.
\newblock In \emph{Proceedings of the 2019 Conference on Empirical Methods in
  Natural Language Processing and the 9th International Joint Conference on
  Natural Language Processing (EMNLP-IJCNLP)}.

\bibitem[{Carreras and M{\`a}rquez(2005)}]{carreras-marquez-2005-introduction}
Xavier Carreras and Llu{\'\i}s M{\`a}rquez. 2005.
\newblock \href {https://www.aclweb.org/anthology/W05-0620} {Introduction to
  the {C}o{NLL}-2005 shared task: Semantic role labeling}.
\newblock In \emph{Proceedings of the Ninth Conference on Computational Natural
  Language Learning ({C}o{NLL}-2005)}, pages 152--164, Ann Arbor, Michigan.
  Association for Computational Linguistics.

\bibitem[{Chen et~al.(2019)Chen, Lyu, and Titov}]{chen-etal-2019-capturing}
Xinchi Chen, Chunchuan Lyu, and Ivan Titov. 2019.
\newblock \href {https://www.aclweb.org/anthology/D19-1544} {{Capturing
  Argument Interaction in Semantic Role Labeling with Capsule Networks}}.
\newblock In \emph{Proceedings of the 2019 Conference on Empirical Methods in
  Natural Language Processing and the 9th International Joint Conference on
  Natural Language Processing (EMNLP-IJCNLP)}.

\bibitem[{Chomsky(1981)}]{chomsky1981lectures}
N.~Chomsky. 1981.
\newblock \href {https://books.google.com.hk/books?id=l08tpkOOdNQC}
  {\emph{{Lectures on Government and Binding}}}.
\newblock Mouton de Gruyter.

\bibitem[{Devlin et~al.(2019)Devlin, Chang, Lee, and Toutanova}]{Jacobbert}
Jacob Devlin, Ming-Wei Chang, Kenton Lee, and Kristina Toutanova. 2019.
\newblock \href {https://www.aclweb.org/anthology/N19-1423} {{{BERT}:
  Pre-training of Deep Bidirectional Transformers for Language Understanding}}.
\newblock In \emph{Proceedings of the 2019 Conference of the North {A}merican
  Chapter of the Association for Computational Linguistics: Human Language
  Technologies (NAACL:HLT)}.

\bibitem[{Dozat and Manning(2017)}]{Dozat2017Deep}
Timothy Dozat and Christopher~D. Manning. 2017.
\newblock {Deep Biaffine Attention for Neural Dependency Parsing}.
\newblock In \emph{International Conference on Learning Representations 2017
  (ICLR)}.

\bibitem[{Fern{\'a}ndez-Gonz{\'a}lez and
  G{\'o}mez-Rodr{\'\i}guez(2019)}]{Daniel-2019-naacl-left}
Daniel Fern{\'a}ndez-Gonz{\'a}lez and Carlos G{\'o}mez-Rodr{\'\i}guez. 2019.
\newblock \href {https://www.aclweb.org/anthology/N19-1076} {{Left-to-Right
  Dependency Parsing with Pointer Networks}}.
\newblock In \emph{Proceedings of the 2019 Conference of the North {A}merican
  Chapter of the Association for Computational Linguistics: Human Language
  Technologies (NAACL)}.

\bibitem[{Gaddy et~al.(2018)Gaddy, Stern, and Klein}]{Gaddy}
David Gaddy, Mitchell Stern, and Dan Klein. 2018.
\newblock {What's Going On in Neural Constituency Parsers? An Analysis}.
\newblock In \emph{Proceedings of the 2018 Conference of the North American
  Chapter of the Association for Computational Linguistics: Human Language
  Technologies (NAACL: HLT)}.

\bibitem[{Haji{\v{c}} et~al.(2009)Haji{\v{c}}, Ciaramita, Johansson, Kawahara,
  Mart{\'\i}, M{\`a}rquez, Meyers, Nivre, Pad{\'o}, {\v{S}}t{\v{e}}p{\'a}nek,
  Stra{\v{n}}{\'a}k, Surdeanu, Xue, and Zhang}]{hajic-etal-2009-conll}
Jan Haji{\v{c}}, Massimiliano Ciaramita, Richard Johansson, Daisuke Kawahara,
  Maria~Ant{\`o}nia Mart{\'\i}, Llu{\'\i}s M{\`a}rquez, Adam Meyers, Joakim
  Nivre, Sebastian Pad{\'o}, Jan {\v{S}}t{\v{e}}p{\'a}nek, Pavel
  Stra{\v{n}}{\'a}k, Mihai Surdeanu, Nianwen Xue, and Yi~Zhang. 2009.
\newblock \href {https://www.aclweb.org/anthology/W09-1201} {The {C}o{NLL}-2009
  shared task: Syntactic and semantic dependencies in multiple languages}.
\newblock In \emph{Proceedings of the Thirteenth Conference on Computational
  Natural Language Learning ({C}o{NLL} 2009): Shared Task}, pages 1--18,
  Boulder, Colorado. Association for Computational Linguistics.

\bibitem[{He et~al.(2018{\natexlab{a}})He, Lee, Levy, and
  Zettlemoyer}]{he-etal-2018-jointly}
Luheng He, Kenton Lee, Omer Levy, and Luke Zettlemoyer. 2018{\natexlab{a}}.
\newblock \href {https://doi.org/10.18653/v1/P18-2058} {Jointly predicting
  predicates and arguments in neural semantic role labeling}.
\newblock In \emph{Proceedings of the 56th Annual Meeting of the Association
  for Computational Linguistics (Volume 2: Short Papers)}, pages 364--369,
  Melbourne, Australia. Association for Computational Linguistics.

\bibitem[{He et~al.(2017)He, Lee, Lewis, and Zettlemoyer}]{he-etal-2017-deep}
Luheng He, Kenton Lee, Mike Lewis, and Luke Zettlemoyer. 2017.
\newblock \href {https://doi.org/10.18653/v1/P17-1044} {Deep semantic role
  labeling: What works and what{'}s next}.
\newblock In \emph{Proceedings of the 55th Annual Meeting of the Association
  for Computational Linguistics (Volume 1: Long Papers)}, pages 473--483,
  Vancouver, Canada. Association for Computational Linguistics.

\bibitem[{He et~al.(2019)He, Li, and Zhao}]{he-etal-2019-syntax}
Shexia He, Zuchao Li, and Hai Zhao. 2019.
\newblock \href {https://www.aclweb.org/anthology/D19-1538} {{Syntax-aware
  Multilingual Semantic Role Labeling}}.
\newblock In \emph{Proceedings of the 2019 Conference on Empirical Methods in
  Natural Language Processing and the 9th International Joint Conference on
  Natural Language Processing (EMNLP-IJCNLP)}.

\bibitem[{He et~al.(2018{\natexlab{b}})He, Li, Zhao, and
  Bai}]{he-etal-2018-syntax}
Shexia He, Zuchao Li, Hai Zhao, and Hongxiao Bai. 2018{\natexlab{b}}.
\newblock \href {https://doi.org/10.18653/v1/P18-1192} {Syntax for semantic
  role labeling, to be, or not to be}.
\newblock In \emph{Proceedings of the 56th Annual Meeting of the Association
  for Computational Linguistics (Volume 1: Long Papers)}, pages 2061--2071,
  Melbourne, Australia. Association for Computational Linguistics.

\bibitem[{Henderson et~al.(2013)Henderson, Merlo, Titov, and
  Musillo}]{henderson-etal-2013-multilingual}
James Henderson, Paola Merlo, Ivan Titov, and Gabriele Musillo. 2013.
\newblock \href {https://doi.org/10.1162/COLI_a_00158} {Multilingual joint
  parsing of syntactic and semantic dependencies with a latent variable model}.
\newblock \emph{Computational Linguistics}, 39(4):949--998.

\bibitem[{Ji et~al.(2019)Ji, Wu, and Lan}]{ji-etal-2019-graph}
Tao Ji, Yuanbin Wu, and Man Lan. 2019.
\newblock \href {https://doi.org/10.18653/v1/P19-1237} {Graph-based dependency
  parsing with graph neural networks}.
\newblock In \emph{Proceedings of the 57th Annual Meeting of the Association
  for Computational Linguistics}, pages 2475--2485, Florence, Italy.
  Association for Computational Linguistics.

\bibitem[{Johansson and Nugues(2008)}]{johansson-nugues-2008-dependency}
Richard Johansson and Pierre Nugues. 2008.
\newblock \href {https://www.aclweb.org/anthology/D08-1008} {Dependency-based
  semantic role labeling of {P}rop{B}ank}.
\newblock In \emph{Proceedings of the 2008 Conference on Empirical Methods in
  Natural Language Processing}, pages 69--78, Honolulu, Hawaii. Association for
  Computational Linguistics.

\bibitem[{Kasai et~al.(2019)Kasai, Friedman, Frank, Radev, and
  Rambow}]{kasai-etal-2019-syntax}
Jungo Kasai, Dan Friedman, Robert Frank, Dragomir Radev, and Owen Rambow. 2019.
\newblock \href {https://doi.org/10.18653/v1/N19-1075} {Syntax-aware neural
  semantic role labeling with supertags}.
\newblock In \emph{Proceedings of the 2019 Conference of the North {A}merican
  Chapter of the Association for Computational Linguistics: Human Language
  Technologies, Volume 1 (Long and Short Papers)}, pages 701--709, Minneapolis,
  Minnesota. Association for Computational Linguistics.

\bibitem[{Kitaev et~al.(2019)Kitaev, Cao, and Klein}]{kitaev2018multilingual}
Nikita Kitaev, Steven Cao, and Dan Klein. 2019.
\newblock \href {https://www.aclweb.org/anthology/P19-1340} {{Multilingual
  Constituency Parsing with Self-Attention and Pre-Training}}.
\newblock In \emph{Proceedings of the 57th Annual Meeting of the Association
  for Computational Linguistics (ACL)}.

\bibitem[{Kitaev and Klein(2018)}]{Kitaev-2018-SelfAttentive}
Nikita Kitaev and Dan Klein. 2018.
\newblock {Constituency Parsing with a Self-Attentive Encoder}.
\newblock In \emph{Proceedings of the 56th Annual Meeting of the Association
  for Computational Linguistics (ACL)}.

\bibitem[{Lewis et~al.(2015)Lewis, He, and Zettlemoyer}]{lewis-etal-2015-joint}
Mike Lewis, Luheng He, and Luke Zettlemoyer. 2015.
\newblock \href {https://doi.org/10.18653/v1/D15-1169} {Joint {A}* {CCG}
  parsing and semantic role labelling}.
\newblock In \emph{Proceedings of the 2015 Conference on Empirical Methods in
  Natural Language Processing}, pages 1444--1454, Lisbon, Portugal. Association
  for Computational Linguistics.

\bibitem[{Li et~al.(2019{\natexlab{a}})Li, Zhang, Jia, and
  Zhao}]{li-etal-2019-gan}
Pengshuai Li, Xinsong Zhang, Weijia Jia, and Hai Zhao. 2019{\natexlab{a}}.
\newblock \href {https://doi.org/10.18653/v1/N19-1307} {{GAN} driven
  semi-distant supervision for relation extraction}.
\newblock In \emph{Proceedings of the 2019 Conference of the North {A}merican
  Chapter of the Association for Computational Linguistics: Human Language
  Technologies, Volume 1 (Long and Short Papers)}, pages 3026--3035,
  Minneapolis, Minnesota. Association for Computational Linguistics.

\bibitem[{Li et~al.(2018{\natexlab{a}})Li, Cai, He, and
  Zhao}]{li-etal-2018-seq2seq}
Zuchao Li, Jiaxun Cai, Shexia He, and Hai Zhao. 2018{\natexlab{a}}.
\newblock \href {https://www.aclweb.org/anthology/C18-1271} {Seq2seq dependency
  parsing}.
\newblock In \emph{Proceedings of the 27th International Conference on
  Computational Linguistics}, pages 3203--3214, Santa Fe, New Mexico, USA.
  Association for Computational Linguistics.

\bibitem[{Li et~al.(2018{\natexlab{b}})Li, He, Cai, Zhang, Zhao, Liu, Li, and
  Si}]{li-etal-2018-unified}
Zuchao Li, Shexia He, Jiaxun Cai, Zhuosheng Zhang, Hai Zhao, Gongshen Liu,
  Linlin Li, and Luo Si. 2018{\natexlab{b}}.
\newblock \href {https://doi.org/10.18653/v1/D18-1262} {A unified syntax-aware
  framework for semantic role labeling}.
\newblock In \emph{Proceedings of the 2018 Conference on Empirical Methods in
  Natural Language Processing}, pages 2401--2411, Brussels, Belgium.
  Association for Computational Linguistics.

\bibitem[{Li et~al.(2019{\natexlab{b}})Li, He, Zhao, Zhang, Zhang, Zhou, and
  Zhou}]{Li-aaai-19}
Zuchao Li, Shexia He, Hai Zhao, Yiqing Zhang, Zhuosheng Zhang, Xi~Zhou, and
  Xiang Zhou. 2019{\natexlab{b}}.
\newblock Dependency or span, end-to-end uniform semantic role labeling.
\newblock In \emph{The Thirty-Third AAAI Conference on Artificial Intelligence
  (AAAI)}.

\bibitem[{Li et~al.(2020)Li, Wang, Chen, Utiyama, Sumita, Zhang, and
  Zhao}]{Li2020Data-dependent}
Zuchao Li, Rui Wang, Kehai Chen, Masso Utiyama, Eiichiro Sumita, Zhuosheng
  Zhang, and Hai Zhao. 2020.
\newblock \href {https://openreview.net/forum?id=S1efxTVYDr} {Data-dependent
  gaussian prior objective for language generation}.
\newblock In \emph{International Conference on Learning Representations}.

\bibitem[{Ling et~al.(2015)Ling, Dyer, Black, Trancoso, Fermandez, Amir,
  Marujo, and Lu{\'\i}s}]{ling-etal-2015-finding}
Wang Ling, Chris Dyer, Alan~W Black, Isabel Trancoso, Ram{\'o}n Fermandez,
  Silvio Amir, Lu{\'\i}s Marujo, and Tiago Lu{\'\i}s. 2015.
\newblock \href {https://doi.org/10.18653/v1/D15-1176} {Finding function in
  form: Compositional character models for open vocabulary word
  representation}.
\newblock In \emph{Proceedings of the 2015 Conference on Empirical Methods in
  Natural Language Processing}, pages 1520--1530, Lisbon, Portugal. Association
  for Computational Linguistics.

\bibitem[{Luo et~al.(2020{\natexlab{a}})Luo, Xiao, and
  Zhao}]{luo2019hierarchical}
Ying Luo, Fengshun Xiao, and Hai Zhao. 2020{\natexlab{a}}.
\newblock {Hierarchical Contextualized Representation for Named Entity
  Recognition}.
\newblock In \emph{The Thirty-Fourth AAAI Conference on Artificial Intelligence
  (AAAI)}.

\bibitem[{Luo and Zhao(2020)}]{luo-zhao-2020-bipartite}
Ying Luo and Hai Zhao. 2020.
\newblock \href {https://www.aclweb.org/anthology/2020.acl-main.571}
  {{Bipartite Flat-Graph Network for Nested Named Entity Recognition}}.
\newblock In \emph{Proceedings of the 58th Annual Meeting of the Association
  for Computational Linguistics (ACL)}.

\bibitem[{Luo et~al.(2020{\natexlab{b}})Luo, Zhao, and Zhan}]{luo2019named}
Ying Luo, Hai Zhao, and Junlang Zhan. 2020{\natexlab{b}}.
\newblock {Named Entity Recognition Only from Word Embeddings}.
\newblock In \emph{Proceedings of the 2020 Conference on Empirical Methods in
  Natural Language Processing (EMNLP)}.

\bibitem[{Lyu et~al.(2019)Lyu, Cohen, and Titov}]{lyu-etal-2019-semantic}
Chunchuan Lyu, Shay~B. Cohen, and Ivan Titov. 2019.
\newblock \href {https://www.aclweb.org/anthology/D19-1099} {{Semantic Role
  Labeling with Iterative Structure Refinement}}.
\newblock In \emph{Proceedings of the 2019 Conference on Empirical Methods in
  Natural Language Processing and the 9th International Joint Conference on
  Natural Language Processing (EMNLP-IJCNLP)}.

\bibitem[{Ma et~al.(2018)Ma, Hu, Liu, Peng, Neubig, and Hovy}]{Ma2018Stack}
Xuezhe Ma, Zecong Hu, Jingzhou Liu, Nanyun Peng, Graham Neubig, and Eduard
  Hovy. 2018.
\newblock {Stack-Pointer Networks for Dependency Parsing}.
\newblock In \emph{Proceedings of the 56th Annual Meeting of the Association
  for Computational Linguistics (ACL)}.

\bibitem[{Marcheggiani et~al.(2017)Marcheggiani, Frolov, and
  Titov}]{marcheggiani-etal-2017-simple}
Diego Marcheggiani, Anton Frolov, and Ivan Titov. 2017.
\newblock \href {https://doi.org/10.18653/v1/K17-1041} {A simple and accurate
  syntax-agnostic neural model for dependency-based semantic role labeling}.
\newblock In \emph{Proceedings of the 21st Conference on Computational Natural
  Language Learning ({C}o{NLL} 2017)}, pages 411--420, Vancouver, Canada.
  Association for Computational Linguistics.

\bibitem[{Marcheggiani and Titov(2017)}]{marcheggiani-titov-2017-encoding}
Diego Marcheggiani and Ivan Titov. 2017.
\newblock \href {https://doi.org/10.18653/v1/D17-1159} {Encoding sentences with
  graph convolutional networks for semantic role labeling}.
\newblock In \emph{Proceedings of the 2017 Conference on Empirical Methods in
  Natural Language Processing}, pages 1506--1515, Copenhagen, Denmark.
  Association for Computational Linguistics.

\bibitem[{Marcus et~al.(1993)Marcus, Santorini, and
  Marcinkiewicz}]{MarcusJ93-2004}
Mitchell~P. Marcus, Beatrice Santorini, and Mary~Ann Marcinkiewicz. 1993.
\newblock \href {http://aclweb.org/anthology/J93-2004} {{Building a Large
  Annotated Corpus of English: The Penn Treebank}}.
\newblock \emph{Computational Linguistics}, 19(2).

\bibitem[{de~Marneffe et~al.(2006)de~Marneffe, MacCartney, and
  Manning}]{Marieffe06generatingtyped}
Marie-Catherine de~Marneffe, Bill MacCartney, and Christopher~D. Manning. 2006.
\newblock \href {http://www.lrec-conf.org/proceedings/lrec2006/pdf/440_pdf.pdf}
  {{Generating Typed Dependency Parses from Phrase Structure Parses}}.
\newblock In \emph{Proceedings of the Fifth International Conference on
  Language Resources and Evaluation ({LREC}{'}06)}.

\bibitem[{Ouchi et~al.(2018)Ouchi, Shindo, and
  Matsumoto}]{ouchi-etal-2018-span}
Hiroki Ouchi, Hiroyuki Shindo, and Yuji Matsumoto. 2018.
\newblock \href {https://doi.org/10.18653/v1/D18-1191} {A span selection model
  for semantic role labeling}.
\newblock In \emph{Proceedings of the 2018 Conference on Empirical Methods in
  Natural Language Processing}, pages 1630--1642, Brussels, Belgium.
  Association for Computational Linguistics.

\bibitem[{Palmer et~al.(2005)Palmer, Gildea, and
  Kingsbury}]{palmer-etal-2005-proposition}
Martha Palmer, Daniel Gildea, and Paul Kingsbury. 2005.
\newblock \href {https://doi.org/10.1162/0891201053630264} {The proposition
  bank: An annotated corpus of semantic roles}.
\newblock \emph{Computational Linguistics}, 31(1):71--106.

\bibitem[{Pennington et~al.(2014)Pennington, Socher, and
  Manning}]{PenningtonD14-1162}
Jeffrey Pennington, Richard Socher, and Christopher Manning. 2014.
\newblock {Glove: Global Vectors for Word Representation}.
\newblock In \emph{Proceedings of the 2014 Conference on Empirical Methods in
  Natural Language Processing (EMNLP)}.

\bibitem[{Pollard and Sag(1994)}]{pollard1994head}
Carl Pollard and Ivan~A Sag. 1994.
\newblock \emph{Head-Driven Phrase Structure Grammar.}
\newblock University of Chicago Press.

\bibitem[{Punyakanok et~al.(2008)Punyakanok, Roth, and
  Yih}]{punyakanok-2008-importance}
Vasin Punyakanok, Dan Roth, and Wen-tau Yih. 2008.
\newblock \href {https://doi.org/10.1162/coli.2008.34.2.257} {The importance of
  syntactic parsing and inference in semantic role labeling}.
\newblock \emph{Computational Linguistics}, 34(2).

\bibitem[{Roth and Lapata(2016)}]{roth-lapata-2016-neural}
Michael Roth and Mirella Lapata. 2016.
\newblock \href {https://doi.org/10.18653/v1/P16-1113} {Neural semantic role
  labeling with dependency path embeddings}.
\newblock In \emph{Proceedings of the 54th Annual Meeting of the Association
  for Computational Linguistics (Volume 1: Long Papers)}, pages 1192--1202,
  Berlin, Germany. Association for Computational Linguistics.

\bibitem[{Schabes et~al.(1988)Schabes, Abeille, and Joshi}]{SCHABESC88-2121}
Yves Schabes, Anne Abeille, and Aravind~K. Joshi. 1988.
\newblock \href {http://aclweb.org/anthology/C88-2121} {Parsing strategies with
  'lexicalized' grammars: Application to tree adjoining grammars}.
\newblock In \emph{Coling Budapest 1988 Volume 2: International Conference on
  Computational Linguistics (COLING)}.

\bibitem[{Shi et~al.(2016)Shi, Teng, and Zhang}]{shi-etal-2016-exploiting}
Peng Shi, Zhiyang Teng, and Yue Zhang. 2016.
\newblock \href {https://doi.org/10.18653/v1/D16-1098} {Exploiting mutual
  benefits between syntax and semantic roles using neural network}.
\newblock In \emph{Proceedings of the 2016 Conference on Empirical Methods in
  Natural Language Processing}, pages 968--974, Austin, Texas. Association for
  Computational Linguistics.

\bibitem[{Stern et~al.(2017)Stern, Fried, and Klein}]{SternD17b}
Mitchell Stern, Daniel Fried, and Dan Klein. 2017.
\newblock \href {https://doi.org/10.18653/v1/D17-1178} {{Effective Inference
  for Generative Neural Parsing}}.
\newblock In \emph{Proceedings of the 2017 Conference on Empirical Methods in
  Natural Language Processing (EMNLP)}, pages 1695--1700.

\bibitem[{Strubell et~al.(2018)Strubell, Verga, Andor, Weiss, and
  McCallum}]{strubell-etal-2018-linguistically}
Emma Strubell, Patrick Verga, Daniel Andor, David Weiss, and Andrew McCallum.
  2018.
\newblock \href {https://doi.org/10.18653/v1/D18-1548} {Linguistically-informed
  self-attention for semantic role labeling}.
\newblock In \emph{Proceedings of the 2018 Conference on Empirical Methods in
  Natural Language Processing}, pages 5027--5038, Brussels, Belgium.
  Association for Computational Linguistics.

\bibitem[{Swayamdipta et~al.(2016)Swayamdipta, Ballesteros, Dyer, and
  Smith}]{swayamdipta-etal-2016-greedy}
Swabha Swayamdipta, Miguel Ballesteros, Chris Dyer, and Noah~A. Smith. 2016.
\newblock \href {https://doi.org/10.18653/v1/K16-1019} {Greedy, joint
  syntactic-semantic parsing with stack {LSTM}s}.
\newblock In \emph{Proceedings of The 20th {SIGNLL} Conference on Computational
  Natural Language Learning}, pages 187--197, Berlin, Germany. Association for
  Computational Linguistics.

\bibitem[{Tan et~al.(2017)Tan, Wang, Xie, Chen, and Shi}]{Tan-Deep-Semantic}
Zhixing Tan, Mingxuan Wang, Jun Xie, Yidong Chen, and Xiaodong Shi. 2017.
\newblock \href {http://arxiv.org/abs/1712.01586} {{Deep Semantic Role Labeling
  with Self-Attention}}.
\newblock \emph{CoRR}, abs/1712.01586.

\bibitem[{Toutanova et~al.(2003)Toutanova, Klein, Manning, and
  Singer}]{Toutanova:2003}
Kristina Toutanova, Dan Klein, Christopher~D. Manning, and Yoram Singer. 2003.
\newblock {Feature-rich Part-of-speech Tagging with a Cyclic Dependency
  Network}.
\newblock In \emph{Proceedings of the 2003 Conference of the North American
  Chapter of the Association for Computational Linguistics on Human Language
  Technology (NAACL)}.

\bibitem[{Vaswani et~al.()Vaswani, Shazeer, Parmar, Uszkoreit, Jones, Gomez,
  Kaiser, and Polosukhin}]{Vaswani17}
Ashish Vaswani, Noam Shazeer, Niki Parmar, Jakob Uszkoreit, Llion Jones,
  Aidan~N Gomez, \L~ukasz Kaiser, and Illia Polosukhin.
\newblock {Attention is All you Need}.
\newblock In \emph{Advances in Neural Information Processing Systems (NIPS)}.

\bibitem[{Yang et~al.(2019)Yang, Dai, Yang, Carbonell, Salakhutdinov, and
  Le}]{XLNet-Zhilin-2019}
Zhilin Yang, Zihang Dai, Yiming Yang, Jaime~G. Carbonell, Ruslan Salakhutdinov,
  and Quoc~V. Le. 2019.
\newblock \href {http://arxiv.org/abs/1906.08237} {Xlnet: Generalized
  autoregressive pretraining for language understanding}.
\newblock In \emph{Thirty-Third Annual Conference on Neural Information
  Processing Systems (NeurIPS)}.

\bibitem[{Zhang et~al.(2018{\natexlab{a}})Zhang, Wang, Utiyama, Sumita, and
  Zhao}]{zhang-etal-2018-exploring}
Zhisong Zhang, Rui Wang, Masao Utiyama, Eiichiro Sumita, and Hai Zhao.
  2018{\natexlab{a}}.
\newblock \href {https://doi.org/10.18653/v1/D18-1511} {Exploring recombination
  for efficient decoding of neural machine translation}.
\newblock In \emph{Proceedings of the 2018 Conference on Empirical Methods in
  Natural Language Processing}, pages 4785--4790, Brussels, Belgium.
  Association for Computational Linguistics.

\bibitem[{Zhang et~al.(2019)Zhang, Huang, and Zhao}]{zhang-etal-2019-open}
Zhuosheng Zhang, Yafang Huang, and Hai Zhao. 2019.
\newblock \href {https://doi.org/10.18653/v1/P19-1154} {Open vocabulary
  learning for neural {C}hinese pinyin {IME}}.
\newblock In \emph{Proceedings of the 57th Annual Meeting of the Association
  for Computational Linguistics}, pages 1584--1594, Florence, Italy.
  Association for Computational Linguistics.

\bibitem[{Zhang et~al.(2018{\natexlab{b}})Zhang, Li, Zhu, Zhao, and
  Liu}]{zhang-etal-2018-modeling}
Zhuosheng Zhang, Jiangtong Li, Pengfei Zhu, Hai Zhao, and Gongshen Liu.
  2018{\natexlab{b}}.
\newblock \href {https://www.aclweb.org/anthology/C18-1317} {Modeling
  multi-turn conversation with deep utterance aggregation}.
\newblock In \emph{Proceedings of the 27th International Conference on
  Computational Linguistics}, pages 3740--3752, Santa Fe, New Mexico, USA.
  Association for Computational Linguistics.

\bibitem[{Zhao et~al.(2009{\natexlab{a}})Zhao, Chen, Kazama, Uchimoto, and
  Torisawa}]{zhao-etal-2009-multilingual-dependency}
Hai Zhao, Wenliang Chen, Jun{'}ichi Kazama, Kiyotaka Uchimoto, and Kentaro
  Torisawa. 2009{\natexlab{a}}.
\newblock \href {https://www.aclweb.org/anthology/W09-1209} {Multilingual
  dependency learning: Exploiting rich features for tagging syntactic and
  semantic dependencies}.
\newblock In \emph{Proceedings of the Thirteenth Conference on Computational
  Natural Language Learning ({C}o{NLL} 2009): Shared Task}, pages 61--66,
  Boulder, Colorado. Association for Computational Linguistics.

\bibitem[{Zhao et~al.(2009{\natexlab{b}})Zhao, Chen, and
  Kit}]{zhao-etal-2009-semantic}
Hai Zhao, Wenliang Chen, and Chunyu Kit. 2009{\natexlab{b}}.
\newblock \href {https://www.aclweb.org/anthology/D09-1004} {Semantic
  dependency parsing of {N}om{B}ank and {P}rop{B}ank: An efficient integrated
  approach via a large-scale feature selection}.
\newblock In \emph{Proceedings of the 2009 Conference on Empirical Methods in
  Natural Language Processing}, pages 30--39, Singapore. Association for
  Computational Linguistics.

\bibitem[{Zhao and Kit(2008)}]{zhao-kit-2008-parsing}
Hai Zhao and Chunyu Kit. 2008.
\newblock \href {https://www.aclweb.org/anthology/W08-2127} {Parsing syntactic
  and semantic dependencies with two single-stage maximum entropy models}.
\newblock In \emph{{C}o{NLL} 2008: Proceedings of the Twelfth Conference on
  Computational Natural Language Learning}, pages 203--207, Manchester,
  England. Coling 2008 Organizing Committee.

\bibitem[{Zhao et~al.(2013)Zhao, Zhang, and Kit}]{Zhao_2013}
Hai Zhao, Xiaotian Zhang, and Chunyu Kit. 2013.
\newblock \href {https://doi.org/10.1613/jair.3717} {Integrative semantic
  dependency parsing via efficient large-scale feature selection}.
\newblock \emph{Journal of Artificial Intelligence Research}, 46:203–233.

\bibitem[{Zhou and Zhao(2019)}]{zhou-zhao-2019-head}
Junru Zhou and Hai Zhao. 2019.
\newblock \href {https://doi.org/10.18653/v1/P19-1230} {Head-driven phrase
  structure grammar parsing on {P}enn treebank}.
\newblock In \emph{Proceedings of the 57th Annual Meeting of the Association
  for Computational Linguistics}, pages 2396--2408, Florence, Italy.
  Association for Computational Linguistics.

\end{thebibliography}
\bibliographystyle{acl_natbib}

\end{document}